\pdfoutput=1
\documentclass{article}
\usepackage{iclr2027_conference,times}
\newif\ifanonymous
\anonymousfalse

\usepackage{amsmath,amsfonts,bm}

\def\eqref#1{equation~\ref{#1}}

\def\1{\bm{1}}

\def\vc{{\bm{c}}}

\def\ve{{\bm{e}}}

\DeclareMathAlphabet{\mathsfit}{\encodingdefault}{\sfdefault}{m}{sl}
\SetMathAlphabet{\mathsfit}{bold}{\encodingdefault}{\sfdefault}{bx}{n}

\newcommand{\E}{\mathbb{E}}

\usepackage{float}
\floatstyle{ruled}
\newfloat{protocol}{tbp}{lop}
\floatname{protocol}{Protocol}
\floatstyle{plain}
\usepackage{hyperref}
\hypersetup{hidelinks,pdfauthor={},pdftitle={RAISE: Diagnosing Acquisition Collapse in Costly LLM Signals}}
\usepackage[protrusion=false]{microtype}
\usepackage{url,booktabs,tabularx,listings,amsmath,amssymb,graphicx,multirow,xcolor,placeins}
\lstdefinestyle{prompt}{basicstyle=\ttfamily\scriptsize,breaklines=true,
  breakindent=0pt,columns=fullflexible,frame=single,framesep=3pt,
  xleftmargin=3pt,xrightmargin=3pt,showstringspaces=false,
  postbreak=\mbox{\textcolor{gray}{$\hookrightarrow$}\space}}
\title{RAISE: Diagnosing Acquisition Collapse\\in Costly LLM Signals}
\newcommand{\rewsnr}{\rho}
\newcommand{\rhostar}{\rho^{\star}}
\newcommand{\acqcost}{\lambda}

\iclrfinalcopy
\author{%
\begin{tabular}[t]{@{}p{0.46\textwidth}p{0.46\textwidth}@{}}
Ying Yuan\textsuperscript{*,\ensuremath{\dagger}} & Yu Wang\textsuperscript{*}\\
\mdseries University of California, San Diego & \mdseries Independent Researcher\\[2.2ex]
Yize Cheng & Xuyang Wu\\
\mdseries University of Maryland, College Park & \mdseries Independent Researcher
\end{tabular}}
\begin{document}
\maketitle
\lhead{}
\renewcommand{\headrulewidth}{0pt}
\noindent{\footnotesize \textsuperscript{*}Equal contribution. \quad
\textsuperscript{\ensuremath{\dagger}}Project lead and corresponding author:
\texttt{yingyuan238@gmail.com}.}
\begin{abstract}
Large language models (LLMs) are increasingly used as costly, on-demand
components in real systems, but calling them indiscriminately can waste
substantial compute, latency, and serving budget. The key deployment question is
therefore not only whether an LLM helps on average, but when it is worth calling.
We identify a common failure mode, which we call \emph{acquisition collapse}: an
LLM signal can appear useful in aggregate or post hoc, yet still provide too
little before-call information to support reliable selective use. We introduce
\textbf{RAISE} (Reward-SNR Actionability in Signal Evaluation), a pre-routing
diagnostic framework for testing whether available evidence supports selective use
before committing to a routing strategy. We instantiate RAISE with Structured
Hypothesis Embeddings (SHE), a frozen-LLM intent signal for recommendation using
one LLM call per user, and evaluate it through controlled, retrospective, and
fresh-cohort studies and a prospective offline pilot whose audit decisions are frozen
before independent outcomes are revealed. Across these settings,
predictable incremental benefit, not average lift alone, distinguishes settings
with recoverable selective value; deployment additionally depends on cost and
operational constraints. Seemingly strong oracle or subgroup gains can disappear
under independent evaluation. More broadly,
RAISE reframes costly inference as an information-acquisition problem: before
paying for an expensive model, tool, sensor, or measurement, first test whether
its value is predictable at decision time. This principle motivates cost-aware
acquisition in settings ranging from agent tool use and stronger-model
consultation to robotic sensing and clinical decision pipelines.
\end{abstract}

\section{Introduction}
\label{sec:intro}
Large language models (LLMs) are increasingly used as optional components inside
larger AI systems: recommenders can call an LLM to infer user intent, agents can
invoke tools or stronger models, and predictive systems can acquire additional
semantic information when
needed~\citep{yao2023react,xi2023kar,ren2024rlmrec,wei2024llmrec}.
These capabilities come with nontrivial compute, latency, and serving costs.
Recent work therefore studies how to route queries across models or selectively
invoke expensive components under a limited
budget~\citep{chen2024frugalgpt,ong2024routellm,dekoninck2025cascade}. At
deployment scale, the question is no longer simply \textbf{whether an LLM helps,
but which calls are worth paying for}.

However, evidence that a costly signal is useful does not necessarily show that
it can be used selectively. Average-use evaluation asks whether performance
improves when the signal is acquired broadly. Routing asks a harder question:
\textbf{can information available before the call predict where that improvement
will occur?} A signal may produce positive average lift, or appear highly
valuable for examples selected after their outcomes are observed, while still
offering too little predictable structure to support a reliable acquisition
policy. Noise and post-selection can further exaggerate apparent subgroup
value~\citep{cawley2010selection,bokelmann2024uplift,bastani2026falsepromises}.

\begin{figure}[t]
\centering
\includegraphics[width=\textwidth]{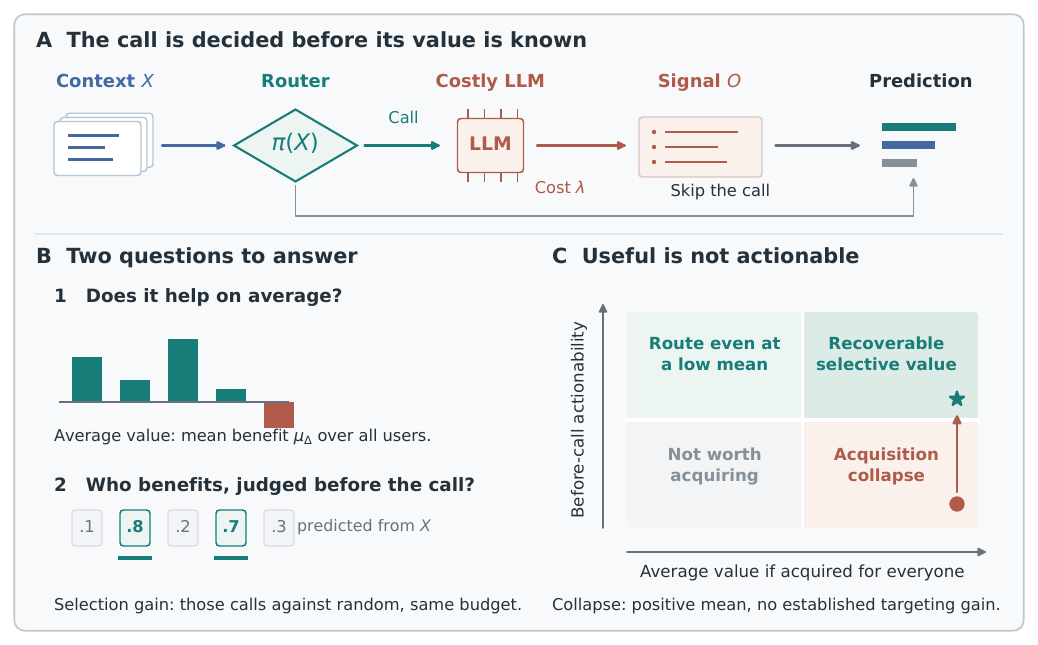}
\caption{\textbf{Acquisition collapse: positive average value without an
established positive held-out selection gain.}
\textbf{A:} From cheap $X$, the router must estimate the incremental
value of costly signal $O$ before paying for it.
\textbf{B:} Average value asks whether acquisition helps overall; before-call
predictability asks where it helps. Selection gain $G_b$ measures selective
value against random acquisition at the same budget.
\textbf{C:} These two axes define four regimes (schematic). Acquisition
collapse is lower right: positive average value without an established positive
held-out selection gain under the evaluated features, policies, budget, and
sample size. Cost is omitted: recoverable selective value can still have negative
net utility (Eq.~\ref{eq:utility}).}
\label{fig:workflow}
\end{figure}

We call this failure mode \textbf{acquisition collapse}: a costly signal appears
valuable under aggregate or post hoc evaluation, yet its value does not
translate into reliable held-out decisions about when to acquire it
(Figure~\ref{fig:workflow}). The distinction is fundamental because acquisition
decisions must be made \emph{before} the expensive observation is available. A
signal can therefore be useful without being actionable.

To understand and diagnose acquisition collapse, this paper addresses two
questions. \textbf{(Q1) How can we tell whether a costly signal is selectively
actionable?} We introduce \textbf{RAISE} (Reward-SNR Actionability in Signal
Evaluation), a pre-routing diagnostic framework that separates overall signal
value from value that can be predicted from before-call information. RAISE
combines reward precision, predictable heterogeneity, independently credited
policy value, and acquisition cost into a common audit.
Its purpose is deliberately upstream of router design: before optimizing how to
route an expensive model, we ask whether the available evidence supports routing
at all. \textbf{(Q2) When does evidence support selective acquisition?} We
instantiate RAISE with \textbf{Structured Hypothesis Embeddings}
(SHE), a frozen-LLM intent signal for recommendation requiring one LLM call per
user. We study it through controlled, retrospective, fresh-cohort, and
prospective offline evaluation, with audit decisions frozen before independent
outcomes are scored. Across these settings, \textbf{predictable incremental
benefit, not average lift alone, distinguishes recoverable selective value}.
Apparent oracle or subgroup gains can vanish under independent evaluation;
deployment further requires measured costs and operational validation.

This perspective connects LLM routing to the broader problem of \textbf{active
information acquisition}~\citep{ma2019eddi,li2021gsm}. Tool calls,
stronger-model consultation, robotic sensing, and costly clinical measurements
all require a system to pay for information before knowing its value. RAISE
provides a common way to ask whether that value is predictable enough to justify
selective acquisition.

Our contributions are threefold.
\begin{enumerate}
\item \textbf{Acquisition collapse.} We identify and formalize a failure mode
that separates the usefulness of a costly signal from its actionability under a
limited acquisition budget.
\item \textbf{RAISE.} We introduce a general pre-routing framework for testing
whether before-call information supports reliable selective acquisition before
investing in a routing policy.
\item \textbf{Evidence across controlled and deployment-style settings.} Using
SHE as a concrete costly LLM signal, we show across controlled, retrospective,
fresh-cohort, and prospective studies that selective value tracks predictable
incremental benefit rather than average lift alone.
\end{enumerate}

\section{Acquisition Collapse and the RAISE Audit}
\label{sec:setup}\label{sec:theory}
RAISE turns the introduction's distinction into an audit by separating
two properties of a costly signal: its average value when acquired and whether
its incremental value is predictable before acquisition.

For each example, $X$ is cheap before-call context, $O$ an optional costly
observation, and $R(1),R(0)$ paired offline rewards from the pipeline with
$O$ (augmented) and without it (base). Let $\Delta=R(1)-R(0)$: $\mu_\Delta=\E[\Delta]$ is average value and
$\tau(x)=\E[\Delta\mid X=x]$ is predictable incremental benefit. A policy
$\pi(X)\in\{0,1\}$ sees only $X$; its evaluation rewards are held out from
fitting and selection. These offline contrasts require separate deployment
validation.

\paragraph{Definition (acquisition collapse).}
Fix a cohort, reward construction, before-call features, budget, sample size,
and policy family. We say that \emph{acquisition collapse} occurs in this
setting when a positive average-use effect, or a post hoc oracle gain from
choosing examples after seeing their rewards, is not matched by an established
positive held-out \emph{selection gain} $G_b$ (Eq.~\ref{eq:policygain}) among the
evaluated policies. A confidence interval that crosses zero leaves selection
unresolved, whereas an upper bound below zero supports worse-than-random
selection.

The same policy is the acquisition gate of an agent: $\pi(X)$ decides whether
to acquire $O$ before a fixed downstream rule acts; in our experiments, it
decides whether to generate SHE before ranking.
\label{sec:agentinterface}

\subsection{Measuring selective value}
When only a fraction $b$ of examples can receive the signal, random acquisition
spends that budget ignoring $X$, while a budgeted policy $\pi_b(X)$ targets
larger predicted benefit. With a constant per-call cost $\acqcost$ expressed in
reward units, net utility relative to skipping is
\begin{equation}
U(\pi)=\E[\pi(X)\Delta]-\acqcost\E[\pi(X)].
\label{eq:utility}
\end{equation}
At budget $\E[\pi_b]=b$, random acquisition has utility
$b\mu_\Delta-b\acqcost$. We define \emph{selection gain} as the policy's
additional utility over random acquisition at the same budget:
\begin{align}
G_b(\pi_b)
&=U(\pi_b)-U(\mathrm{random}_b)
=\E[(\pi_b(X)-b)\Delta] \nonumber\\
&=\operatorname{Cov}(\pi_b(X),\tau(X)).
\label{eq:policygain}
\end{align}
Thus $G_b>0$ means that the policy preferentially acquires examples whose
conditional benefit exceeds the population average; under constant per-call
cost, cost cancels from $G_b$.

\looseness=-1
Eq.~\ref{eq:policygain} is the population average prescriptive effect (PAPE) at
treatment rate $b$~\citep{imai2023experimental}; $G_b/b$ is the selected-group
effect relative to the population mean, the targeting operator characteristic
of the rank-weighted average treatment effects (RATE)
framework~\citep{yadlowsky2025rate}. We use $G_b$ as the endpoint.

\paragraph{Selective value is invariant to a uniform benefit shift.}
Adding the same constant $c$ to every incremental reward leaves a fixed
rate-$b$ policy's gain unchanged:
$\E[(\pi_b-b)(\Delta+c)]=G_b$. Balanced $X\in\{-1,+1\}$ with
$\tau(X)=aX$, $a>0$, has zero mean but $G_{.5}=a/2$ when acquiring at $X=+1$;
constant $\tau(X)=\mu>0$ gives $G_b=0$ for every rate-$b$ policy. Average
value and actionability therefore vary independently.

\subsection{The RAISE audit}
\label{sec:audit}
\looseness=-1
RAISE asks four questions in order: What is compared? Is the reward signal
measurable? Is benefit predictable before acquisition? Does held-out policy
value justify its cost? An audit sample that acquires $O$ for every example,
observing both rewards, answers them while keeping apart the measurements, the
frozen rule that reads them, and the router.

\begin{protocol}[H]
\caption{RAISE: from pre-routing diagnostics to deployment evidence}
\label{prot:raise}
\small
\textbf{Inputs:} paired rewards $R(0),R(1)$; before-call covariates $X$;
budget $b$; benefit target $\epsilon$; cost model; policies.
\begin{enumerate}
\item \textbf{Fix the comparison.} Record the cohort, sampling unit, reward
pipeline, feature allowlist, splits, tuning rules, and planned inference, and
keep evaluation rewards outside fitting and selection.
\item \textbf{Characterize the channel.} Measure signed average value, reward
signal-to-noise ratio (reward-SNR), precision, and feature-predictable
heterogeneity against appropriate nulls and positive controls; freeze
prospective decisions before outcomes.
\item \textbf{Credit selection independently.} Fit policies within training data,
evaluate $G_b$ on held-out rewards, state the scope of intervals, and separate
perturbation references from tests with a justified null.
\item \textbf{Compare acquisition choices.} Report evidence for $G_b>0$,
attainment of $\epsilon$, and net utility against random acquisition,
acquire-all, and skip, preserving unresolved comparisons.
\end{enumerate}
\textbf{Output:} diagnostics, intervals, and supported decisions, tied to the
cohort, reward pipeline, budget, and cost.
\end{protocol}

Step~2 first asks whether average value is measurable. Reward-SNR
$\rewsnr=|\mu_\Delta|/\sigma_\Delta$ is the standardized mean effect. Together
with sample size, it determines the precision of the sample mean. We report the sign of $\mu_\Delta$ and
$N_{\min}\approx(2.8/\rewsnr)^2$, the sample size at which a two-sided
$\alpha=.05$ test has 80\% planned power for the mean. $N_{\min}$ calibrates
mean-effect precision (Appendix~\ref{app:power_sens}). To ask whether that value
can be targeted, we use heterogeneous treatment effect (HTE) diagnostics:
out-of-fold (OOF) score--reward correlation, prediction $R^2$, and group average
treatment effects (GATES)-style contrasts~\citep{chernozhukov2018generic};
matched positive controls with injected signal establish sensitivity.
Permutation tests are valid when they recompute the statistic under a justified
transformation~\citep{ojala2010permutation,hemerik2018exact}; refitting on
disrupted rewards and crediting on real ones is only a stability reference. These diagnostics indicate whether
targeting structure exists; held-out $G_b$ measures whether a policy recovers it.

Step~4 turns evidence into decisions, which require both a practical target
and a cost model. Against
target $\epsilon>0$, interval $[L,H]$ is \emph{confirmed-useful} if
$L>\epsilon$, \emph{below-target} if $H\le\epsilon$, and \emph{unresolved}
otherwise. Crossing $\epsilon$ after clearing zero means positive gain but
unresolved target attainment. By Eq.~\ref{eq:utility}, net value at budget $b$
is $b(\mu_\Delta-\lambda)+G_b$, so a positive gain may still be uneconomic;
variable costs require comparing expected expense. With uniform cost and no overhead, selection beats acquire-all and
skip only if its net value exceeds
$\max\{\mu_\Delta-\lambda,0\}$ (Appendix~\ref{app:costfrontier}).

\section{Experimental Instantiation and Study Design}
\label{sec:method}\label{sec:sheaudit}
RAISE needs paired outcomes with and without the costly signal. Recommendation
provides them because base and LLM-augmented pipelines can score the same users.
We use Structured Hypothesis Embeddings (SHE) as the costly observation $O$.
Given a user history, a frozen LLM produces $K=3$ intent hypotheses $h_k$,
confidence scores $\gamma_k$, and indices that point to supporting history
events. With frozen hypothesis embeddings $\ve_k=\phi(h_k)$ and candidate
embedding $\vc$, we summarize the best confidence-weighted match as
\begin{equation}
f_B^{\max}(\vc)=\max_{k=1,\ldots,K}\gamma_k\cos(\vc,\ve_k).
\label{eq:she}
\end{equation}
This score is one coordinate of a four-feature SHE block; the original
Amazon-Beauty analysis used a legacy variant (Appendix~\ref{app:she}). A late-fusion ranker
concatenates that block with the base recommender score; Appendix~\ref{app:she}
details the full block, encoders, and generator dependence. We then test
selective value in four stages, each removing a source of optimistic bias
(Table~\ref{tab:studymap}): a synthetic controlled study
(Section~\ref{sec:validate}) followed by three real-data studies
(Section~\ref{sec:results}).

\begin{table}[H]
\caption{\textbf{Four stages, four questions.} Each stage strengthens a
distinct separation.}
\label{tab:studymap}
\centering\footnotesize
\setlength{\tabcolsep}{3pt}
\begin{tabularx}{\linewidth}{@{}>{\raggedright\arraybackslash}p{.18\linewidth}
 >{\raggedright\arraybackslash}p{.34\linewidth}
 >{\raggedright\arraybackslash}X@{}}
\toprule
Stage & Key separation & Question \\
\midrule
Controlled & Aligned/permuted features; fixed rewards &
Does predictability move $G_b$ without changing the mean? \\
Original & Cross-fitted rewards; independent policy credit &
Do oracle or subgroup gains transfer? \\
Fresh MIND & Frozen upstream models; OOF routers &
Does held-out selection emerge? \\
Prospective & Disjoint audit/train/evaluation; frozen decisions &
Do frozen predictions transfer? \\
\bottomrule
\end{tabularx}
\end{table}

We begin retrospectively. On the original Microsoft News Dataset
(MIND)~\citep{wu2020mind} ($N=1263$) and the Amazon Reviews 2023 All\_Beauty
cohort (Amazon-Beauty; $N=650$)~\citep{hou2024bridging}, cross-fitted rewards
replace \emph{self-credit}, in which the same rewards both choose and score a
selection: each evaluation unit is scored by reward models fit without its
fold. Fresh MIND then adds $N=4954$ unseen users while keeping
prior-cohort-fitted latent semantic analysis (LSA) representations, gated
recurrent unit (GRU) and self-attentive sequential recommendation (SASRec)
backbones~\citep{hidasi2016gru4rec,kang2018sasrec}, and ranking heads frozen;
routers are trained and credited out of fold. The stages use different
populations and reward pipelines (Appendices~\ref{app:cohorts}, \ref{app:data},
\ref{app:hparams}, and~\ref{app:robust}).

Those studies remain retrospective. The final gap is temporal: can the audit
commit before outcomes are known? The \emph{prospective offline pilot} tests this
on new
MIND and Amazon Reviews 2023 Beauty and Personal Care
users~\citep{hou2024bridging}. It is prospective in chronology, because audit
decisions and policy selections are frozen before evaluation rewards are scored,
but offline in evaluation, because it uses historical records and ranking
outcomes rather than a live intervention. It is a pilot because its four
conditions (GRU and SASRec on each domain) can test transfer but are too few to
estimate the framework's error rates. MIND GRU is the sole pre-specified primary endpoint;
MIND SASRec and both Amazon conditions are secondary. Each domain assigns 800, 1,600,
and 1,600 users to audit, router training, and evaluation. Backbones share users
and generated signals; historical reward models remain frozen. We evaluate the
fitted policy in all four conditions regardless of the audit decision. The
router sees ten before-call features: candidate and history counts, four
history--candidate similarity summaries, two normalized backbone-score summaries,
and sparse-history and multi-intent indicators; SHE outputs enter only the
ranker (Appendices~\ref{app:cohorts}, \ref{app:prospective},
and~\ref{app:prospectiveconfig}, including GPT-5.5 provenance limits).

Uncertainty follows each stage's sampling unit. In the prospective offline pilot, the
endpoint is $G_{.30}$ over budget-matched random acquisition at a 30\% call
budget. Ridge policies fit on router users and select 480 of the 1,600 evaluation
users before reward scoring. Ten thousand shared-user bootstrap draws give
fixed-policy intervals conditional on fitted models and selections; we compare
them with the pre-specified target $\epsilon=.005$ in normalized discounted
cumulative gain at rank 10 (NDCG@10)~\citep{jarvelin2002cumulated}. Fresh MIND uses five-fold OOF
policies with refit sensitivities, whereas controlled intervals span draw seeds.

\section{Controlled Evidence: Predictability Creates Selective Value}
\label{sec:validate}\label{sec:scan}
Before turning to real cohorts, a synthetic Gaussian study tests the prediction
of Section~\ref{sec:setup}: selection gain follows predictable heterogeneity,
not average benefit. Each user has eight standard-Gaussian before-call
features; true benefit is a common mean plus a term along one fixed feature
direction, and independent Gaussian noise produces training and evaluation
rewards. Three Gaussian cases vary these two ingredients. In the
constant-positive case, everyone has the same positive benefit, so acquisition
helps on average but offers no better target. The positive-mean heterogeneous
case adds predictable user-level differences. The zero-mean heterogeneous case
removes only the common mean shift, preserving those differences around zero
(Appendix~\ref{app:controlled}).

Within each case, the aligned arm gives routers the true features, while the
permuted arm shuffles feature rows across users; both arms share rewards, users,
and folds, so only feature--benefit alignment changes. If predictability matters,
permutation should erase $G_b$, while constant-positive should show none.
Policies train on one noise realization and are credited on an independent one
(\emph{noisy credit}); \emph{latent credit} instead scores them against true
benefit. Settings were fixed on three development draws; the primary comparison
is aligned-minus-permuted gain across 20 separate paired validation draws.

\begin{figure}[t]
\centering
\includegraphics[width=\textwidth]{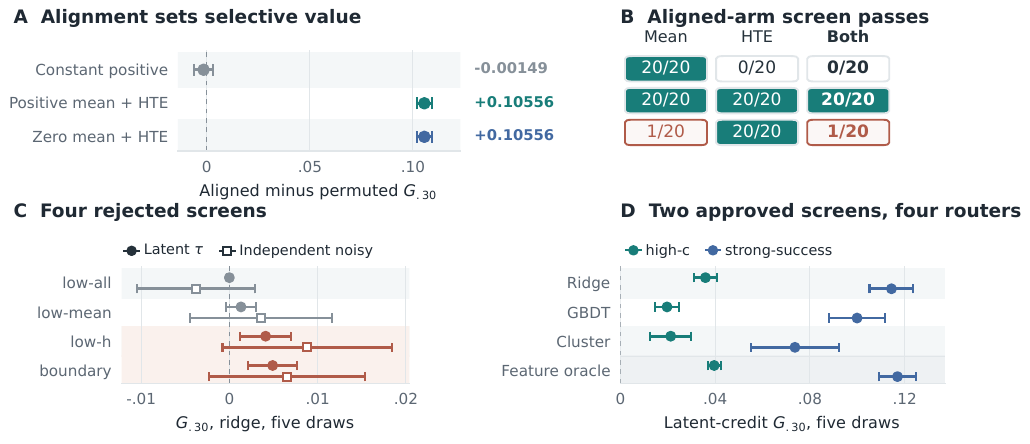}
\caption{\textbf{Predictable incremental benefit drives selective value.}
Values are in synthetic reward units; each panel has its own $x$ scale.
\textbf{A:} Aligned and permuted features share rewards; only alignment
changes. Points show independent-noisy-credit $G_b$ with paired Student-$t$ 95\%
intervals over 20 draws; the heterogeneous cases differ by a constant mean
shift, so their gains coincide.
\textbf{B:} Aligned-arm pass counts: requiring both mean and HTE screens
withholds approval from the zero-mean case in 19/20 draws.
\textbf{C:} Frozen six-case scan (five draws); two shaded false negatives are
positive under latent but unresolved under noisy credit.
\textbf{D:} Every router from the two approved cases; the feature oracle is
an upper-bound reference.}
\label{fig:mechanism}
\end{figure}

That is what we observe. Relative to permuted features, aligned features add
$+0.10556$ noisy-credit selection gain in both heterogeneous cases, with 95\%
interval $[+0.10179,+0.10933]$; constant-positive has average value but gives
$-0.00149$ $[-0.00595,+0.00298]$ aligned-minus-permuted selection-gain contrast
(Figure~\ref{fig:mechanism}). Identical gains confirm invariance to the mean
shift.

\noindent\begin{minipage}{\linewidth}
The zero-mean case exposes the cost of requiring average value and heterogeneity
together. A conjunctive screen approves only when both a positive-mean test and
an HTE test pass. The zero-mean aligned arm passes the HTE screen in 20/20
draws but the positive-mean screen in only 1/20, so the conjunctive screen
withholds approval in 19/20 despite recoverable selective value. This miss is
structural: larger samples sharpen inference around zero while the aligned
heterogeneity persists.
\end{minipage}

\looseness=-1
A six-case scan checks the pattern beyond this construction: six further
Gaussian cases, five draws each, screened by a conjunctive rule frozen before
outcomes. Its two approved cases
retain independent-noisy-credit gains of $0.03624$ $[0.02548,0.04699]$ and
$0.11705$ $[0.10160,0.13250]$; all three learned routers (ridge, gradient-boosted
trees, and a cluster router) recover value under both latent and noisy credit
(Appendix~\ref{app:controlled}).

\section{Real-Data Evidence: From Collapse to Prospective Transfer}
\label{sec:results}
\begin{figure}[t]
\centering
\includegraphics[width=\textwidth]{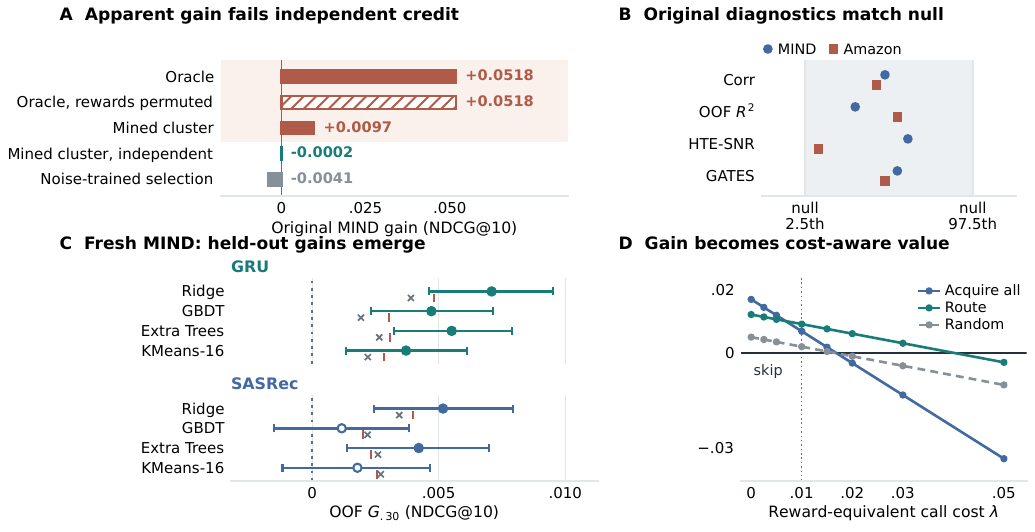}
\caption{\textbf{Independent credit separates apparent gain from held-out
selective value.} Cohorts and reward pipelines differ; all intervals are 95\%.
\textbf{A:} On original MIND ($N=1263$), a self-credited oracle survives reward
permutation, while a mined cluster loses its gain under independent credit;
shading marks reused sorting rewards.
\textbf{B:} All eight original MIND/Amazon diagnostics lie inside their own
permutation-null bands.
\textbf{C:} On fresh MIND ($N=4954$, $b=.30$), six of eight OOF policies have
positive policy-by-policy (marginal) intervals; ticks and crosses are
uncalibrated perturbation references.
\textbf{D:} GRU-ridge routing beats acquire-all only beyond the dotted cost;
skip wins at high cost.}
\label{fig:evidence}
\end{figure}

We now test whether the controlled pattern survives the three real-data stages
of Table~\ref{tab:studymap}. Figure~\ref{fig:evidence} tracks the first two and
their cost implications.

\subsection{Original cohorts: apparent gain fails independent credit}
\label{sec:acquisition}
We start with self-credit on original MIND. A self-credited oracle, which
selects the examples with the highest realized rewards, still gains
$+0.0518$ after reward permutation because one reward realization both ranks
and credits examples (Figure~\ref{fig:evidence}A). A cluster mined for high
reward gains $+0.0097$ when the same rewards credit it but falls to $-0.0002$ under independent credit;
a selector trained on noise and credited on real rewards yields $-0.0041$.
Positive OOF selection remains unresolved across the evaluated policy family:
apparent value fails independent credit, the signature of acquisition collapse.

Could a different router exploit heterogeneity missed by these policies? On
MIND, OOF score--reward correlation is $-0.0181$, $R^2=-0.0102$, and the
grouped top-minus-bottom contrast is $+0.00254$. HTE-SNR, the standard deviation
of OOF predictions divided by residual standard deviation, is $0.0829$. Every
statistic, like its Amazon counterpart, falls inside its permutation-null band
(Figure~\ref{fig:evidence}B); a sensitivity analysis with reward models fit on
disjoint users leaves $R^2$ and the grouped contrast unresolved. Positive controls that inject known
effects into eight k-means clusters (KMeans-8) are first recovered at
cluster-SNR, between- over within-cluster reward standard deviation, of $.20$ on
MIND and $.35$ on Amazon, above the real $.0799$ and $.0628$, so sensitivity is limited at these
sizes (Appendix~\ref{app:hte}). Two explanations remain: no predictable signal
or too little data.

\subsection{Fresh MIND: held-out selection emerges}
\label{sec:fresh}\label{sec:scaling}
Fresh MIND probes the data-limitation explanation, although its cohort and
reward pipeline also change. With $N=4954$ unseen users,
frozen upstream models, and OOF router credit, selective value becomes
detectable. Six of eight policies, all reported in this exploratory stage, have
fixed-policy intervals entirely above zero; these marginal intervals are not
adjusted across policies (Figure~\ref{fig:evidence}C). Ridge gains are $0.00709$
$[0.00460,0.00951]$ for
GRU and $0.00517$ $[0.00244,0.00794]$ for SASRec. Average lift stays
positive across historical cohorts of increasing size, while positive policy
intervals emerge only at larger sizes (Figure~\ref{fig:f4scaling}). Ten
acquisition-split refits
and five reward-pipeline conditions keep 59/60 and 28/30 nominal intervals above
zero as correlated stability summaries (Appendix~\ref{app:f4}). Average-benefit
detection and selective-policy learning therefore emerge on different
sample-size scales. The 1,000-draw permuted-reward and matched-noise refits in
Figure~\ref{fig:evidence}C lack an exchangeability justification for calibrated
null tests (Appendix~\ref{app:inferencegap}).

With 1,600 evaluation users fixed and router-training size varied from 100 to
3,354, both maximum-size ridge intervals exclude zero. The curves are
nonmonotonic, however, and both maximum-minus-500-user contrasts cross zero.
These data therefore do not establish a routing-onset threshold
(Figure~\ref{fig:fixedlearning}; Appendix~\ref{app:learning}).

When is selection worth paying for? At approximately 30\% calls, the
selected increment
$V_b=\E[\pi_b\Delta]=b\mu_\Delta+G_b$ retains 71.86\% $[58.30\%,89.22\%]$ of the
always-acquire increment with GRU and 68.51\% $[48.59\%,99.12\%]$ with SASRec,
under joint fixed-policy bootstrap. Acquire-all is preferable while calls
are nearly free, routing becomes preferable at nontrivial cost, and skipping
dominates when acquisition is expensive (Figure~\ref{fig:evidence}D). Costs are
in reward-equivalent units (Appendix~\ref{app:costfrontier}).

\subsection{Prospective offline pilot: frozen decisions, open rule comparison}
\label{sec:prospective}
\begin{figure}[t]
\centering
\includegraphics[width=\textwidth]{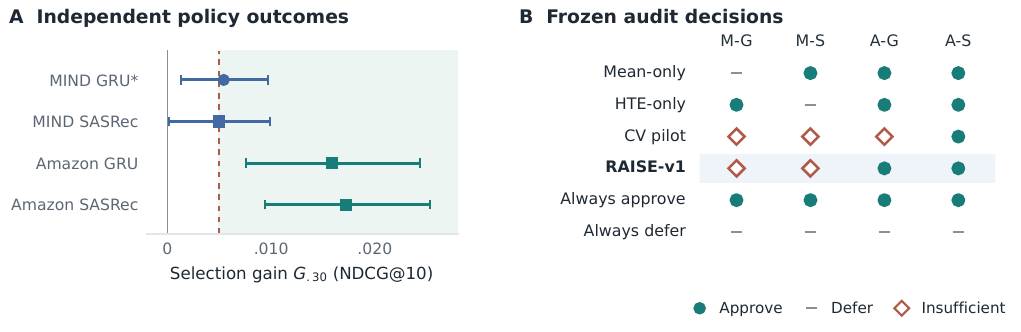}
\caption{\textbf{Frozen audit decisions and independent policy outcomes answer
different questions.}
\textbf{A:} Four prospective ridge endpoints at $b=.30$ with 95\% fixed-policy
intervals and target $\epsilon=.005$ dashed. MIND GRU is the sole primary: MIND
clears zero but not the target; correlated Amazon secondaries clear it.
\textbf{B:} Six rules frozen before outcomes; M/A and G/S denote dataset and
backbone. Intervals condition on fitted policies and locked selections.}
\label{fig:prospective}
\end{figure}

Fresh MIND establishes held-out value, but not whether an audit would predict it
before the final outcomes. We froze six rules or references. Mean-only approves
a significantly positive audit mean; HTE-only approves when a permutation-tested
HTE statistic passes. RAISE-v1 uses an exact three-way agreement rule: approve
when both components approve, defer when both defer, and record insufficient
evidence when they disagree or required evidence is missing. The broader
framework treats these dimensions separately. A small cross-validated (CV) router audit
(``CV pilot'' in Figure~\ref{fig:prospective}) checks its gain interval against
$\epsilon$; always-approve and always-defer are constant references. All four data-dependent rules share one 800-user audit sample
(Appendix~\ref{app:prospective}).

On MIND, both fixed-policy intervals exclude zero: the primary GRU endpoint gains
$G_{.30}=0.00543$ (95\% interval $[0.00129,0.00974]$) and SASRec $0.00497$
$[0.00017,0.00986]$ (Figure~\ref{fig:prospective}). Both cross the pre-specified
target $\epsilon=.005$, leaving practical usefulness unresolved. RAISE-v1 had
recorded insufficient evidence for both. The component screens disagree: mean-only
approves SASRec, whereas HTE-only approves GRU; against the target, these
approvals remain unclassified.

On Amazon, both correlated secondary endpoints exceed the target: GRU gains
$0.01587$ $[0.00762,0.02437]$ and SASRec $0.01718$
$[0.00937,0.02528]$. RAISE-v1, mean-only, HTE-only, and always-approve had all
approved both; the CV pilot approved SASRec and abstained on GRU.

The pilot establishes that decisions frozen before outcomes can be checked
against later outcomes, not that any rule is superior. It does not establish a
performance ordering among the audit rules: several rules make the same
approvals on the two confirmed-useful Amazon conditions, while both MIND
conditions remain unresolved relative to the target. With no resolved
below-target condition, a broader prospective study is needed to compare false
approvals and quantify RAISE-v1's incremental value.

\section{Related Work}
\label{sec:related}
\ifanonymous
\else
\paragraph{LLM routing and cost-aware inference.}
Recent work reduces LLM inference cost through cascades, model routing, and
selective invocation. FrugalGPT learns cascades across models with different
costs and capabilities~\citep{chen2024frugalgpt}, RouteLLM learns query-level
routing from preference data~\citep{ong2024routellm}, and later work unifies
the two~\citep{dekoninck2025cascade}. RouterBench and LLMRouterBench standardize
evaluation of routing systems~\citep{hu2024routerbench,li2026llmrouterbench}.
These methods develop how to route once a routing problem and model pool have
been specified.

\paragraph{Realizability diagnostics for routing.}
Recent diagnostics also ask whether hindsight routing opportunity is
realizable. \citet{shihab2026opportunity} distinguish outcome-oracle,
signal-restricted, and held-out learned-router gains, together with
selection-valid inference. \citet{chen2026routinggap} separate reproducible
specialist advantage from single-draw label noise. These analyses narrow any
simple divide between work that asks how to route and work that asks whether
routing is warranted. RAISE is complementary: its decision is whether to buy an
optional costly observation, and its estimands use paired acquire/skip rewards.
The audit places average value, before-call predictable heterogeneity,
independently credited selection gain, and acquisition cost in one evidence
sequence before checking frozen decisions on new cohorts.

\paragraph{Active information acquisition, deferral, and selective prediction.}
Active feature acquisition chooses which measurements to obtain from partially
observed inputs~\citep{ma2019eddi,li2021gsm}, while learning to defer and
selective prediction decide when to rely on an alternative predictor or
expert~\citep{madras2018defer,mozannar2020defer,geifman2017selective}; conformal
routing constrains violations against an expensive-model
reference~\citep{uddin2026conformal}. RAISE does not propose another acquisition
policy. It tests whether before-call information supports selective acquisition
under a stated budget and cost model, so the audit can precede routing,
deferral, or acquisition algorithms.

\paragraph{Heterogeneous effects and policy evaluation.}
The distinction between average and selective value connects to heterogeneous
treatment-effect estimation and policy learning. Effect estimation and
decision quality are not the same objective~\citep{fernandezloria2022causal}.
Empirical welfare maximization and policy learning target decisions
directly~\citep{kitagawa2018welfare,athey2021policy,bastani2025inferenceaware}.
PAPE and RATE quantify improvement from targeting over non-targeted
allocation~\citep{imai2023experimental,yadlowsky2025rate}, while GATES-style
analyses inspect predictable effect variation~\citep{chernozhukov2018generic}.
RAISE adopts these ideas for costly signal acquisition, using held-out
selection value to ask whether before-call context identifies where an optional
observation is most useful. Its contribution is not a new HTE estimator, but a
pre-routing audit that combines mean value, predictable heterogeneity,
independent policy credit, and acquisition cost.
\fi

\paragraph{Diagnosing hidden failures in LLM systems.}
Aggregate performance can also conceal the dependence that deployment needs.
RAGEN-2 diagnoses reasoning collapse in agentic RL by showing that aggregate
indicators can miss the relationship between inputs and useful reasoning
behavior~\citep{wang2026ragen2}. RAISE studies a different dependence: whether
information available before acquisition predicts the incremental value of a
costly signal. The shared lesson: aggregate success is insufficient when deployment
depends on a finer-grained relationship. Acquisition collapse names
this failure for costly acquisition and gives a way to test for it before
routing is deployed.

\section{Conclusion and Limitations}
\label{sec:discussion}
\paragraph{Conclusion.}
\ifanonymous
\else
Costly LLM signals create a deployment problem that average performance alone
cannot answer: even when an LLM helps, a system still needs to know when the
call is worth paying for. We identify \textbf{acquisition collapse}, in which a
signal appears valuable in aggregate or post hoc but provides too little
before-call information for reliable selective acquisition. We introduce
\textbf{RAISE} (Reward-SNR Actionability in Signal Evaluation) to diagnose this
gap by separating overall signal value from value that is predictable before
acquisition. Using \textbf{Structured Hypothesis Embeddings} (SHE) as a
frozen-LLM signal for recommendation, we evaluate the audit through controlled,
retrospective, fresh-cohort, and prospective offline studies. Independent
credit explains why apparent oracle or subgroup gains can disappear, while the
fresh and prospective stages show when held-out selection can recover value.
Across these settings, \textbf{predictable incremental benefit, not average
lift alone, distinguishes recoverable selective value}; deployment additionally
depends on measured acquisition costs and operational validation. More broadly,
the results recast costly LLM invocation as an information-acquisition problem:
before building a router around an expensive model, tool, sensor, or
measurement, first test whether its value is predictable at decision time.
\fi

\paragraph{Limitations.}
Our current evidence focuses on single-step recommendation and one LLM-signal
design, motivating broader validation across models, tasks, and domains. The
prospective offline pilot validates frozen-before-outcome auditing and positive
held-out selection, but does not yet establish when RAISE-v1 adds
value beyond simpler screens. Production studies should incorporate measured
monetary cost, latency, routing overhead, and acquisition failures rather than
reward-equivalent costs alone. Extending RAISE to agents, robotics, and other
sequential settings will additionally require history-dependent rewards,
trajectory-level evaluation, and accounting for how one acquisition changes
later decisions and future calls (Appendix~\ref{app:agentinterface}).
The original Amazon study uses review-derived action proxies and
next-reviewed-item ranking rather than logged browsing or purchase events;
Appendix~\ref{app:data} states the mapping and the resulting prompt mismatch.
\label{sec:mainend}

\subsection*{AI Use Statement}
The experimental method uses a frozen LLM to generate structured intent
hypotheses from histories, as described in Section~\ref{sec:method} and
Appendix~\ref{app:she}. AI tools also assisted with literature search.

\subsubsection*{Reproducibility Statement}
The real-data studies use public recommendation datasets. We document reward
definitions, data roles, frozen rules, prompts, and uncertainty procedures in
the main text and appendices. Preserved experiment packages contain source
snapshots, manifests, cached outputs, and offline reconstruction scripts;
the prospective endpoint and bootstrap calculations have been independently
reconstructed. This verification covers cached numerical evidence, with
upstream reproduction and historical provenance limits described in
Appendix~\ref{app:provenance}. These packages are preserved internally and are
not currently public. Code will be released upon acceptance.

\subsubsection*{Author Contributions}
Ying Yuan led the project as corresponding author, conceived the
research question and core thesis, formulated the costly LLM signal-acquisition
problem and reward-SNR diagnostic framing, designed the Structured Hypothesis
Embeddings method, and led the experimental pipeline, including data processing,
hypothesis generation, backbone and acquisition studies, positive controls,
robustness analyses, result interpretation, revision strategy, figure/table
design, and manuscript writing. Yu Wang contributed to experiment execution,
figure preparation, technical discussions informing the second version,
critical manuscript review, and writing. Yize Cheng and Xuyang Wu contributed
substantive discussions and recommendations for the second version, helping
refine the manuscript's framing, interpretation, and presentation.

\subsubsection*{Acknowledgments}
We thank Daiwei Chen for helpful discussion and suggestions. This work uses only
publicly available datasets (MIND, REES46, and Amazon review data); no proprietary data
were used.

\bibliographystyle{iclr2027_conference}
\bibliography{references}
\clearpage
\appendix
\section{Detailed Cohort Boundaries}
\label{app:cohorts}
Table~\ref{tab:cohorts} maps each study to its evaluation population and reward
pipeline. The original signal analyses and the follow-up acquisition studies
use different representations, sample sizes, and outcomes. We identify the
cohort and backbone with each SNR estimate so that the comparisons retain
their intended meaning.

\begin{table}[h]
\caption{Analysis-specific cohorts. The controlled screen and its router
replication share the same experiment.
Fresh MIND changes more than sample size relative to the original MIND study.}
\label{tab:cohorts}
\centering\small
\begin{tabularx}{\linewidth}{@{}>{\raggedright\arraybackslash}p{.24\linewidth}
 r >{\raggedright\arraybackslash}X@{}}
\toprule
Analysis & $N$ & Reward and evaluation scope \\
\midrule
Original MIND acquisition & 1,263 & ada-002 content/late-fusion study;
cross-fitted ranking rewards; corrected heterogeneity analysis. \\
Original Amazon-Beauty & 650 & Amazon Reviews 2023 All\_Beauty; local LSA representation;
cross-fitted base and augmented ranking rewards. \\
Original REES46 & 498 & Thin-content session-purchase study;
separate binary reward/probe setting, not fresh-MIND NDCG routing. \\
Original MIND backbone study & 1,411 & LSA-based ordered/unordered
backbones; average-use contrast, not the fresh-MIND test cohort. \\
Controlled scan & 4,000 & Six cases, five draws each;
noisy training reward and latent-effect policy credit. \\
Fresh MIND & 4,954 & One impression per user; prior-cohort-fitted,
frozen LSA, GRU/SASRec, and ranking heads; realized NDCG@10 reward. \\
Fixed-test follow-up & 1,600 & One cached test set; disjoint router
training ladder with 100--3,354 users. \\
Frozen-head MIND sensitivity & 715 & Heterogeneity audit after a separate
476-user reward-head fit and identity-ambiguity quarantine. \\
Prospective MIND & 4,000 & Disjoint 800/1,600/1,600 audit, router,
and evaluation roles; GRU is the sole primary endpoint. \\
Prospective Amazon & 4,000 & New Beauty and Personal Care users;
same role sizes; both backbone outcomes are secondary. \\
\bottomrule
\end{tabularx}
\end{table}

The prior MIND generation artifact contains 1,557 rows, of which 1,549 are
valid. Analysis-specific ranking eligibility produces the smaller legacy
cohorts above. Fresh MIND selects 5,000 distinct new users, with 4,954 valid
generated rows. The delivered selection and overlap manifests record
exclusions of both
the prior analysis cohort and a previously inspected 6,000-row pool. The
follow-up provides source snapshots, split hashes, and chronology records;
some original generation failures and historical embedding artifacts remain
unavailable. Each analysis therefore retains its own eligibility boundary.

For the original analyses, impression/window resampling may not account for
repeated users or overlapping histories. Fresh MIND has one impression per user and
uses user-level resampling. This addresses repeated-user dependence within the
cohort but does not remove dependence induced by overlapping OOF training sets
or shared news environments.

\section{Controlled Screen and Outcome Details}
\label{app:controlled}\label{app:csens}
\paragraph{Controlled-family construction.}
Let $\tau_{\mathrm{lat}}$ be the simulated per-example effect, distinguished
from the conditional mean $\tau(x)=\E[\tau_{\mathrm{lat}}\mid X=x]$.
The scale family separates a predictable component $s(X)$, an unobserved
component $u$, and observation noise $\varepsilon$:
\begin{equation}
\tau_{\mathrm{lat}}=\mu+\sqrt{h}\,\sigma_{\mathrm{het}}s(X)
       +\sqrt{1-h}\,\sigma_{\mathrm{het}}u,
\qquad \Delta_{\mathrm{obs}}=\tau_{\mathrm{lat}}+\varepsilon.
\label{eq:scanmodel}
\end{equation}
Figures and tables use the short label $\tau$ for latent credit.
For independent, centered, unit-variance $s,u,\varepsilon$ and positive
$\mu$, setting $\sigma_{\mathrm{het}}=c\mu$ in Eq.~\ref{eq:scanmodel} gives
$\rho=\mu/\sqrt{1+c^2\mu^2}$. Hence, when $c\rho<1$,
\begin{equation}
\mu=\frac{\rho}{\sqrt{1-c^2\rho^2}},\qquad
\sigma_{\mathrm{sig}}=\sqrt{h}\sigma_{\mathrm{het}}
=\sqrt{h}\frac{c\rho}{\sqrt{1-(c\rho)^2}}.
\label{eq:crho}
\end{equation}
Eq.~\ref{eq:crho} explains how the controlled family varies predictable
signal: its dispersion is tied to its mean. Other acquisition problems can vary
these quantities independently. The resulting scale is useful for constructing
the scan; a learner's finite-sample behavior still depends on the feature
representation, policy class, and estimation procedure.

\paragraph{Corrected six-case scan.}
The controlled scan uses $N=4000$, seeds 10000--10004, and budget $b=0.30$.
The corrected source removes the post-call fields \texttt{b\_conf} and
\texttt{b\_max}; the eight retained covariates are available before acquisition.
The binary screen requires a positive mean above its planning reference and
an HTE test passing either the $R^2$ or grouped-contrast branch. Case-level
PASS requires this conjunction in at least three of five draws. This frozen
six-case rule differs from the three-way prospective agreement rule in
Appendix~\ref{app:prospective}.

Figure~\ref{fig:mechanism}C,D draws these five-draw outcomes: the four rejected
cases under both credit endpoints, and every router fitted in the two approved
cases. All cases and matched-noise controls are retained, and these results
precede the independent-noisy-credit and 20-draw mechanism follow-ups.
Tables~\ref{tab:f1full} and~\ref{tab:f3routers} list the same estimates
numerically.

\begin{table}[h]
\caption{All frozen screen labels and primary ridge outcomes. The
screen uses noisy observed reward; policy credit uses latent $\tau$.
Positive intervals in \texttt{low-h} and \texttt{boundary} are false negatives
of the reported screen and are not relabeled using cost.}
\label{tab:f1full}
\centering\small
\setlength{\tabcolsep}{4pt}
\begin{tabular}{llrrrl}
\toprule
Case & Screen & $\rho/\rho^\star$ & HTE-SNR & $G_{0.30}$ on $\tau$ & 95\% CI \\
\midrule
low-all & FAIL & 0.282 & 0.0420 & $+0.00000$ & $[-0.00022, +0.00022]$ \\
low-mean & FAIL & 0.532 & 0.0438 & $+0.00132$ & $[-0.00035, +0.00299]$ \\
low-h & FAIL & 2.031 & 0.0479 & $+0.00411$ & $[+0.00126, +0.00697]$ \\
boundary & FAIL & 1.032 & 0.0490 & $+0.00491$ & $[+0.00213, +0.00769]$ \\
high-c & PASS & 2.031 & 0.1101 & $+0.03589$ & $[+0.03095, +0.04084]$ \\
strong-success & PASS & 4.034 & 0.3061 & $+0.11447$ & $[+0.10520, +0.12373]$ \\
\bottomrule
\end{tabular}

\end{table}

The mean observed rewards for \texttt{low-all}, \texttt{low-mean},
\texttt{low-h}, \texttt{boundary}, \texttt{high-c}, and
\texttt{strong-success} are respectively $0.01260$, $0.02368$, $0.09043$,
$0.04585$, $0.09150$, and $0.19090$. The corresponding mean per-draw
$N_{\min}$ values are 124,630; 58,189; 1,037; 4,898; 1,036; and 250.
These average nonlinear plug-in estimates across draws, so they need not equal
$(2.8/\bar\rho)^2$ calculated from the displayed mean SNR. The \texttt{high-c}
case targets $\rho/\rho^\star=2$, $h=0.5$, $c=2$; the
\texttt{strong-success} case targets $4$, $1$, and $2$.

\begin{table}[h]
\caption{Multiple routers recover value in the two PASS settings. These are
the same draws used by the screen, not an independent replication.}
\label{tab:f3routers}
\centering\small
\begin{tabular}{llrr}
\toprule
Case & Router & Latent-credit gain & 95\% CI \\
\midrule
high-c & Ridge & $+0.03589$ & $[+0.03095,+0.04084]$ \\
high-c & GBDT & $+0.01966$ & $[+0.01476,+0.02456]$ \\
high-c & Cluster & $+0.02123$ & $[+0.01248,+0.02998]$ \\
strong-success & Ridge & $+0.11447$ & $[+0.10520,+0.12373]$ \\
strong-success & GBDT & $+0.09993$ & $[+0.08820,+0.11166]$ \\
strong-success & Cluster & $+0.07370$ & $[+0.05513,+0.09226]$ \\
\bottomrule
\end{tabular}
\end{table}

Ridge recovers 90.5\% of the feature-oracle value in \texttt{high-c}, with
95\% CI $[83.3\%,97.6\%]$, and 97.8\% in \texttt{strong-success},
$[95.9\%,99.7\%]$. The absolute-utility sensitivity uses
$\lambda=2\max(\overline{\Delta}_{\mathrm{obs}},10^{-4})$, a data-derived
within-case cost. The resulting utilities are $+0.00798$
$[-0.00745,+0.02341]$ and $+0.05674$ $[+0.03812,+0.07535]$,
respectively. These absolute-utility intervals answer a different question
from the budget-matched contrast. All six matched-noise-trained placebo
intervals contain zero; their credit uses observed reward, not latent $\tau$.

\paragraph{Independent noisy credit.}
The follow-up fixes the original policies and generates a new noise
realization, $\Delta_{\mathrm{new}}=\tau_{\mathrm{lat}}+
\varepsilon_{\mathrm{new}}$, for evaluation. The latent effects and users are
unchanged; the independence concerns outcome noise. Table~\ref{tab:noisycredit}
reports all six primary ridge outcomes. Across-draw intervals use a
Student-$t$ critical value with four degrees of freedom. The original
\texttt{low-h} and \texttt{boundary} false negatives remain part of the latent
analysis, while their noisier-credit intervals now cross zero.

\begin{table}[h]
\caption{Primary ridge outcomes under independent noisy credit. Original
latent-credit false negatives are retained as historical classifications;
new uncertainty does not retroactively change the frozen decisions.}
\label{tab:noisycredit}
\centering\small
\begin{tabular}{@{}llrrl@{}}
\toprule
Case & Screen & Latent gain & Noisy gain & Noisy 95\% interval \\
\midrule
low-all & FAIL & $+0.00000$ & $-0.00376$ & $[-0.01042,+0.00290]$ \\
low-mean & FAIL & $+0.00132$ & $+0.00363$ & $[-0.00442,+0.01167]$ \\
low-h & FAIL & $+0.00411$ & $+0.00881$ & $[-0.00077,+0.01839]$ \\
boundary & FAIL & $+0.00491$ & $+0.00655$ & $[-0.00226,+0.01536]$ \\
high-c & PASS & $+0.03589$ & $+0.03624$ & $[+0.02548,+0.04699]$ \\
strong-success & PASS & $+0.11447$ & $+0.11705$ & $[+0.10160,+0.13250]$ \\
\bottomrule
\end{tabular}

\end{table}

\paragraph{Paired alignment intervention.}
The mechanism study freezes three cases after three development draws and
evaluates 20 separate validation draws. Each draw compares an aligned arm
with a feature-permuted arm using the same marginal rewards, noise, and
evaluation users. Each draw contains $N=4000$ units, eight independent standard
Gaussian before-call features, and budget $b=.30$. Let $s(X)$ be the centered,
standardized projection of $X$ onto the direction
$[1,-.8,.6,.4,0,0,0,0]$. The three cases use
$(\mu,\sigma_{\mathrm{het}})=(.1,0),(.1,.3),(0,.3)$ and
\[
\tau=\mu+\sigma_{\mathrm{het}}s(X),\qquad
\Delta_{\mathrm{obs}}=\tau+\varepsilon_{\mathrm{obs}},\qquad
\Delta_{\mathrm{new}}=\tau+\varepsilon_{\mathrm{new}},
\]
where the two noise terms are independent standard Gaussians. They are the
constant-positive, positive-mean heterogeneous, and zero-mean heterogeneous
cases, respectively. The permuted arm reassigns complete $X$ rows while keeping
$\tau$, both reward realizations, users, and folds fixed. The primary contrast is
the within-draw difference in
independent-noisy-credit $G_b$; its interval summarizes 20 paired differences,
not independent arm means. The constant-positive aligned-minus-permuted
contrast is $-0.00149$
$[-0.00595,0.00298]$. Positive-heterogeneous and zero-mean heterogeneous cases
both give $0.10556$ $[0.10179,0.10933]$: their constant mean shift leaves
budget-matched selection unchanged. In the zero-mean case the mean gate passes
1/20 draws, the HTE gate 20/20, and their conjunction 1/20. In the constant
case those counts are 20/20, 0/20, and 0/20. In the positive-heterogeneous case
all three pass 20/20. These synthetic reward units are not NDCG points.

\section{Corrected Heterogeneity Audit on the Original Cohorts}
\label{app:hte}\label{app:poscontrol}
\begin{figure}[h]
\centering
\includegraphics[width=\textwidth]{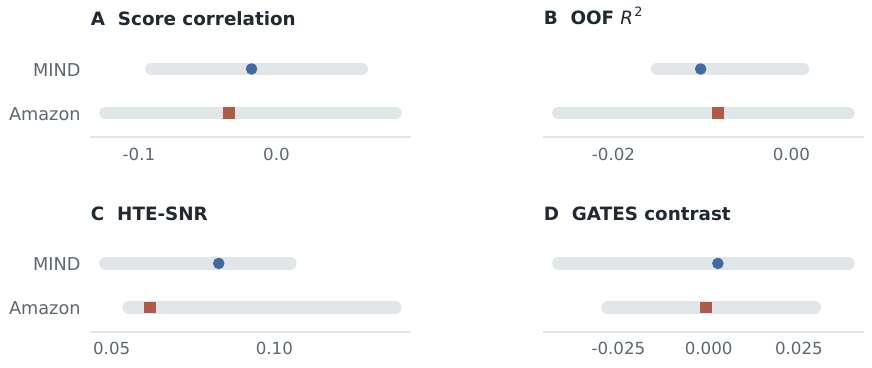}
\caption{\textbf{All eight original-cohort heterogeneity diagnostics lie within
their permutation-null bands.} Points show MIND and Amazon estimates; grey
segments show the reported 95\% permutation-null ranges. These are reference
ranges, not confidence intervals
around the points. Each metric uses its own scale.}
\label{fig:hte}
\end{figure}

The heterogeneity analysis fits an OOF ridge predictor to a kNN-smoothed reward target constructed
within each training fold. GATES-style groups are formed within held-out
folds. The base and augmented rewards themselves use a separate
impression-level cross-fitting partition. This is layered OOF evaluation,
not fully nested reward/heterogeneity cross-fitting. The follow-up below
separates reward-head fitting from the heterogeneity sample.
For OOF predictions $\widehat\Delta$, the reported heterogeneity signal-to-noise
ratio is
$\mathrm{HTE\mbox{-}SNR}=\operatorname{SD}(\widehat\Delta)/
[\operatorname{SD}(\Delta-\widehat\Delta)+10^{-12}]$, using population standard
deviations and equal unit weights. It measures variation in predicted effects
relative to residual reward variation; it is not a treatment-effect estimator
or a group-based cluster-SNR.
Figure~\ref{fig:hte} places each diagnostic on its own scale beside its
permutation-null range, which Figure~\ref{fig:evidence}B rescales away.

\begin{table}[h]
\caption{Corrected heterogeneity diagnostics, transcribed from the frozen
analysis record. The
reported HTE-SNR is not the cluster-SNR used by the KMeans positive control.
The delivered source and cached null draws specify the estimators and their
permutation implementation.}
\label{tab:f2full}
\centering\small
\begin{tabular}{llrrr}
\toprule
Cohort & Diagnostic & Estimate & Permutation-null 95\% band & $p$ \\
\midrule
MIND & Correlation & $-0.0181$ & $[-0.0909, +0.0620]$ & 0.532 \\
MIND & $R^2$ & $-0.0102$ & $[-0.0151, +0.0013]$ & 0.781 \\
MIND & HTE-SNR & $0.0829$ & $[0.0481, 0.1049]$ & 0.259 \\
MIND & GATES spread & $+0.00254$ & $[-0.04163, +0.03869]$ & 0.483 \\
Amazon-Beauty & Correlation & $-0.0345$ & $[-0.1243, +0.0865]$ & 0.607 \\
Amazon-Beauty & $R^2$ & $-0.00824$ & $[-0.02619, +0.00641]$ & 0.348 \\
Amazon-Beauty & HTE-SNR & $0.0618$ & $[0.0552, 0.1372]$ & 0.915 \\
Amazon-Beauty & GATES spread & $-0.00073$ & $[-0.02803, +0.02943]$ & 0.547 \\
\bottomrule
\end{tabular}

\end{table}

Table~\ref{tab:f2full} lists every diagnostic. On MIND, score correlation is
positive in three of five folds, $R^2$ in one of five, and the grouped
top-minus-bottom contrast in two of five.
Amazon-Beauty has negative $R^2$ in all five folds and varying grouped-contrast
signs. The null results support no detected stable heterogeneity for these
tested estimators and samples. They do not establish absence of an effect
under every feature representation or policy class.

The corrected cluster-SNR values are $0.0799$ for MIND and $0.0628$ for
Amazon-Beauty. Here cluster-SNR is the unit-weighted between-cluster standard
deviation of cluster-level effects divided by the unit-weighted within-cluster
residual standard deviation; the matched controls vary this ratio by injecting
cluster effects. KMeans-8
first recovers the matched injected signals at cluster-SNR $0.20$ and $0.35$,
respectively. It measures structure available to this grouping procedure and is
distinct from HTE-SNR. Global mean reward-SNR
describes a different property and is reported separately in
Appendix~\ref{app:power_sens}.

\paragraph{Frozen-head sensitivity.}
The stricter follow-up fits the reward head on 476 users and audits a disjoint
715-user set. Mean reward is $0.014995$, prediction $R^2=-0.018523$
($p=.825$), and the grouped contrast is $-0.020909$ ($p=.781$), using 999
permutations. This sensitivity does not recover detectable heterogeneity.
It has no embedded positive-control sweep; the matched recovery controls
above belong to the original heterogeneity analysis. Identity reconstruction quarantines
45 ambiguous groups involving 99 possible users and checks available
history-prefix mappings. Missing original ada-002 artifacts and generation
logs limit claims about complete historical identity recovery.

\section{Fresh-MIND Protocol and Complete Results}
\label{app:f4}
\paragraph{Reward isolation.}
The LSA encoder, ordered GRU/SASRec models, and base/augmented ranking heads
are trained on the prior cohort before the fresh cohort is scored. The
original fresh-MIND run has one frozen configuration and seed per backbone. It uses five fixed
folds, four router families, and a 30\% budget. $N$ denotes the total OOF
cohort; router training uses approximately four fifths of it. The frozen
analysis record lists 2,000 bootstrap draws, 50 repetitions per refitted placebo family, and ten
matched positive-control repetitions.

Figure~\ref{fig:evidence}C shows all eight policy estimates with their
perturbation references; Table~\ref{tab:f4full} lists the same values.
Appendix~\ref{app:learning} varies router-training size on a fixed test cohort.

\begin{table}[h]
\caption{Full fresh-cohort policy family ($N=4954$, $b=0.30$). The last two
columns are empirical 97.5th percentiles of refitted reward-permutation and
trained-noise policy gains, credited on real held-out rewards. They are not
confidence bounds on the real policy. Intervals condition on fitted OOF
policies and are not adjusted for eight comparisons.}
\label{tab:f4full}
\centering\small
\setlength{\tabcolsep}{3pt}
\begin{tabular}{llrrrr}
\toprule
Backbone & Router & $G_{0.30}$ & 95\% CI & Perm. upper & Noise upper \\
\midrule
GRU & Ridge & $+0.00709$ & $[+0.00460, +0.00951]$ & $+0.00482$ & $+0.00390$ \\
GRU & GBDT & $+0.00471$ & $[+0.00233, +0.00714]$ & $+0.00304$ & $+0.00193$ \\
GRU & Extra Trees & $+0.00551$ & $[+0.00322, +0.00789]$ & $+0.00308$ & $+0.00265$ \\
GRU & KMeans-16 & $+0.00371$ & $[+0.00132, +0.00613]$ & $+0.00283$ & $+0.00220$ \\
\midrule
SASRec & Ridge & $+0.00517$ & $[+0.00244, +0.00794]$ & $+0.00399$ & $+0.00344$ \\
SASRec & GBDT & $+0.00117$ & $[-0.00151, +0.00384]$ & $+0.00200$ & $+0.00219$ \\
SASRec & Extra Trees & $+0.00421$ & $[+0.00138, +0.00698]$ & $+0.00232$ & $+0.00260$ \\
SASRec & KMeans-16 & $+0.00179$ & $[-0.00117, +0.00464]$ & $+0.00257$ & $+0.00271$ \\
\bottomrule
\end{tabular}

\end{table}

\paragraph{Null and positive controls.}
The $6$ prefixes $\times$ $2$ backbones $\times$ $4$ routers $\times$
$2$ placebo types yield 96 reported null summaries. All have intervals
containing zero; the largest absolute mean is $0.00168$. These overlapping
summaries are not 96 independent studies and do not establish calibrated
family-wise error. The subsequent permutation follow-up increases each reference family
to 1,000 repetitions. Its train-on-disrupted, credit-on-real construction
requires a separate validity argument before its exceedance rates can be
interpreted as permutation $p$-values; increasing the repetition count does
not supply that argument.

At the full cohort size, the matched positive-control pipeline reaches at
least 80\% recovery at reported HTE-SNR $0.10$ for all GRU routers and
for SASRec ridge, GBDT, and Extra Trees. SASRec KMeans requires $0.20$.
This recovery criterion concerns a designed heterogeneous signal, not the
80\% mean-test power convention in Eq.~\ref{eq:quickdiag}.

\paragraph{Self-credited oracle.}
The fresh-cohort self-credited oracle gains are $0.04253$ for GRU and
$0.05565$ for SASRec. Because sorting and crediting use the same realized
reward, these quantities contain order-statistic selection effects. They are
not attainable policy ceilings, and the difference from a learned router does
not estimate remaining predictable heterogeneity.

\paragraph{Attrition and eligibility.}
Forty-six selected rows fail the frozen complete-row eligibility rule:
10/11/9/5/11 across the five folds, or 0.92\% overall. Excluded rows have
longer histories on average (21.85 versus 15.66). Retaining only eligible rows
changes the sparse-history share by 0.19 percentage points, which is a limited
distribution check rather than proof that exclusion is ignorable. Results
condition on generation-valid users; a production fallback policy would also
need to include failures.

The cohort construction retains at least one positive candidate per source
impression. This is a label-informed offline slate construction and remains
outside the router feature channel. The delivered source/retained-slate
mapping, before-call feature allowlist, and overlap checks document the
eligible population. Detailed causes for all 46 original generation failures
are not recoverable. One impression per user reduces repeated-user dependence,
but does not make shared training folds independent replications.

\subsection{Inference boundary and refit stability}
\label{app:inferencegap}
The permutation follow-up permutes training rewards, refits OOF scores, and
credits pooled top-budget selections on the original rewards. Let $S(X,y)$
denote the OOF scores and
$g(S,y)$ their credited selection gain. Its reference statistic is
$g(S(X,Py),y)$ for permutation $P$, whereas a full-statistic permutation of
$T(X,y)=g(S(X,y),y)$ evaluates $T(X,Py)$.
These distributions need not coincide. OOF fitting alone also does not make
the pooled selection indicators independent of their credited outcomes:
global reward permutation can move an outcome into a training fold used to
score that user, and pooled thresholds depend on other fitted scores.

An exact counterexample uses ten users in five identical-feature pairs, with
pair members assigned to opposite folds. A one-nearest-neighbor OOF rule
predicts the partner's reward and selects the top three scores. Under an
exchangeable reward null, enumerating the 945 equally likely pairings in the
provided construction yields 75 nominal $5\%$ rejections, or
$5/63\approx7.94\%$. This disproves OOF fitting as a general calibration
argument. It neither estimates the fresh-MIND false-positive rate nor shows that its
observed gains are false. Valid permutation inference requires a stated null,
admissible transformations, and an appropriately recomputed
statistic~\citep{hemerik2018exact}. Formal family-wise significance for the
eight fresh-MIND policies remains unresolved; historical exceedance rates are retained
in the frozen archive as perturbation summaries.

The prospective HTE code uses a different construction: it fits and credits
each statistic on the same permuted reward vector. The train/credit mismatch
above does not transfer to that calculation. Its null assumptions still need
to match the sampling design. Prospective policy intervals likewise condition
on the already locked policy; they do not include repeated pipeline training.

\begin{table}[h]
\caption{Fresh-MIND refit stability for all eight policies. Counts give positive
point estimates and positive marginal 95\% intervals, respectively. The
ten outer-split refits and five pipeline conditions reuse the cohort and
are sensitivity analyses, not independent replications or multiplicity control.}
\label{tab:refitstability}
\centering\small
\begin{tabular}{@{}llrrrrrr@{}}
\toprule
Backbone & Router & Split mean & Gain+ & CI+ & Pipeline mean & Gain+ & CI+ \\
\midrule
GRU & Ridge & $+0.00667$ & 10/10 & 10/10 & $+0.00641$ & 5/5 & 5/5 \\
GRU & GBDT & $+0.00365$ & 10/10 & 10/10 & $+0.00571$ & 5/5 & 5/5 \\
GRU & Extra Trees & $+0.00520$ & 10/10 & 10/10 & $+0.00615$ & 5/5 & 5/5 \\
GRU & KMeans-16 & $+0.00369$ & 10/10 & 10/10 & $+0.00442$ & 5/5 & 4/5 \\
SASRec & Ridge & $+0.00475$ & 10/10 & 10/10 & $+0.00442$ & 5/5 & 4/5 \\
SASRec & GBDT & $+0.00142$ & 8/10 & 1/10 & $+0.00261$ & 5/5 & 2/5 \\
SASRec & Extra Trees & $+0.00344$ & 10/10 & 9/10 & $+0.00527$ & 5/5 & 5/5 \\
SASRec & KMeans-16 & $+0.00186$ & 10/10 & 1/10 & $+0.00339$ & 5/5 & 3/5 \\
\bottomrule
\end{tabular}

\end{table}

Table~\ref{tab:refitstability} reports all eight policies. Five pipeline
conditions comprise the preserved canonical run and four full seed refits. A
same-seed rerun is not bit-identical to the canonical export;
both are retained. Across the six originally positive policies, 60/60
outer-refit estimates and 59/60 intervals are positive; the corresponding
pipeline counts are 30/30 and 28/30. These checks support stability of the
observed pattern under the tested changes without certifying a family-wise test.

\subsection{Training size with a fixed test cohort}
\label{app:learning}
\begin{figure}[h]
\centering
\includegraphics[width=\textwidth]{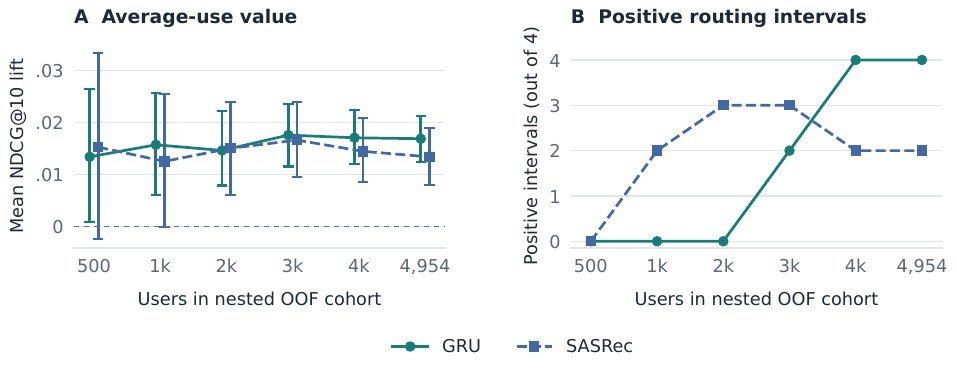}
\caption{\textbf{Historical fresh-MIND scaling is descriptive, not a universal
onset threshold.} \textbf{A:} Average-use mean lift with 95\% bootstrap
intervals. \textbf{B:} Number of the four routers with positive marginal policy
intervals for each backbone, using the interval criterion alone with no placebo
filter. Reward summaries and evaluated users both change across sizes, so the
ladder does not isolate the effect of training sample size.}
\label{fig:f4scaling}
\end{figure}

The original ladder changes the evaluated cohort as $N$ grows
(Figure~\ref{fig:f4scaling}). The fixed-test follow-up instead holds 1,600
cached test users and a 480-user acquisition budget constant, with disjoint
router-training sets of 100, 250, 500, 1,000, 2,000, and 3,354 users.
All 48 backbone/router/size cells and 16 paired training-size contrasts are
retained. These cached test rewards had been inspected previously, so this
is a retrospective training-size sensitivity, not a new confirmatory test.

\begin{figure}[h]
\centering
\includegraphics[width=.92\textwidth]{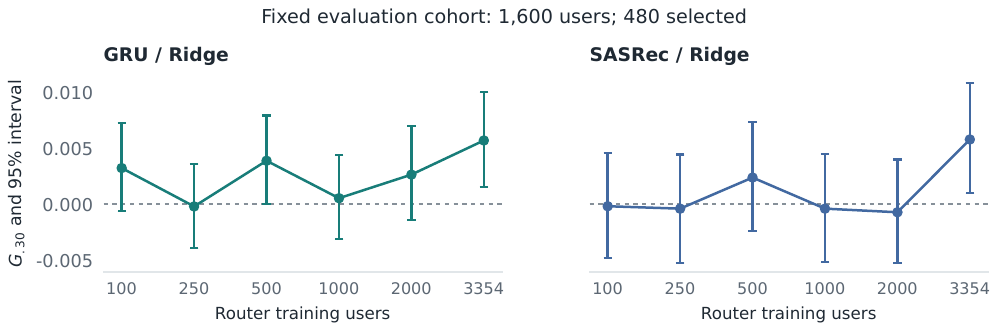}
\caption{\textbf{More router-training users do not yield a resolved monotone
gain.} Fresh-MIND ridge selection gain on a fixed 1,600-user evaluation cohort
with 480 users selected. Points and 95\% fixed-policy intervals use the same
evaluation rewards at every training size. Both maximum-size intervals exceed
zero, but both 3,354-minus-500 contrasts cross zero.}
\label{fig:fixedlearning}
\end{figure}

Figure~\ref{fig:fixedlearning} shows the training-size curve on the fixed
evaluation cohort; Section~\ref{sec:fresh} reads it as evidence against a
universal onset threshold.

At 3,354 training users, ridge gains are $0.005693$
$[0.001545,0.010018]$ for GRU and $0.005784$
$[0.000980,0.010857]$ for SASRec. The paired 3,354-minus-500 differences are
$0.001814$ $[-0.001913,0.005618]$ and $0.003409$
$[-0.001050,0.008004]$, respectively. Pointwise intervals use 10,000 fixed-policy
user bootstrap draws. These comparisons leave the incremental benefit of
additional router-training users unresolved at the tested sizes.

\section{Prospective Offline Pilot: Frozen Decisions and Independent Outcomes}
\label{app:prospective}
The pilot is prospective in chronology but offline in evaluation: it freezes
audit decisions and policy selections before scoring historical ranking
outcomes; it is not a live intervention or A/B test. Each source has 800 audit
users, 1,600 independent
router-training users, and 1,600 evaluation users. GRU and SASRec share
users, generated hypotheses, and role assignments within a source. MIND GRU
is the sole primary endpoint; the other three conditions are secondary.
The Amazon extension adds a second source without changing that designation.

\subsection{Eligibility and information boundaries}
MIND generation yields 4,005 complete users from 4,100 initial requests,
with 95 terminal failures. The first 4,000 complete users populate the roles;
five surplus users and all failures remain archived. Amazon generation yields
4,096 complete users and four terminal failures; the first 4,000 populate the
roles and 96 remain surplus. Both analyses condition on complete generation.
The deployment value of a fallback for failed acquisition is not measured.

The Amazon source is Beauty and Personal Care from Amazon Reviews 2023.
Its historical reward models are fitted on All\_Beauty before fresh
generation. All 631,986 All\_Beauty users are excluded from the fresh source.
Sampling uses a fixed lowest-5\%-hash prefix with 36,350 eligible users;
this is the inspected prefix, not a global eligibility census. The task uses
the same review-derived action mapping as the original Amazon study:
\texttt{purchase} means verified purchase with rating at least four, and every
other review is \texttt{view}; there are no cart actions. It evaluates
next-review relevance proxies with static product metadata and deterministic
hash-selected negatives. MIND retains a positive candidate in each offline
slate. These choices define reproducible offline rewards, with a narrower
interpretation than label-blind production acquisition.

The final router is ridge regression with regularization parameter 1.
Its scaler is fitted on router-training users only. Ten allowed features
are available before the call; post-call SHE values and evaluation rewards
are excluded. Each model ranks by predicted gain and selects
$\operatorname{round}(.30\times1600)=480$ evaluation users. Exact score ties
are ordered by a SHA-256 hash of rule version, setting, user, and impression,
then lexicographically by user and impression. The selection mask is locked
before evaluation rewards are scored. Pre/post evaluation features match exactly
in the delivered checks. Every final policy is evaluated, including policies
whose audit abstains or defers.

\subsection{Generation and router configuration}
\label{app:prospectiveconfig}
Both prospective sources request the deployment alias
\texttt{Shared-gpt-5.5-paygo}, using Scheme B, high reasoning effort, and a
1,536-token completion cap per request. The configuration allows at most
two requests per user, 300 retries shared across the cohort, and 4,400 total
requests for 4,100 initial users. The delivered configuration and launch
records specify the requested service; the exact served-model snapshot and
billing rates are unavailable. Generated hypotheses are cached and shared
across backbones and audit methods. These runs therefore evaluate one
requested generator configuration, not robustness across LLM families.

The prospective reward checkpoints are source-specific and never fit on audit,
router, or evaluation users. MIND reuses the 256-dimensional LSA encoder fitted
on text from its 1,557 prior windows, together with the GRU/SASRec backbones,
15-epoch schedule, and four base/augmented heads fitted on the 1,549 eligible
prior windows, which cover 1,514 unique users (training hash recorded in the
frozen manifest).
Amazon fits one shared 256-dimensional LSA encoder, five OOF models per
backbone, one full GRU and one full SASRec model, and four logistic heads on 646
eligible users from the 650-user All\_Beauty source; no Beauty and Personal Care
user enters fitting. Each cohort is scored only by its matching source and
backbone artifacts. These manifests and checkpoints are preserved internally
but are not currently public.

The router uses the following ten-field allowlist:
\begin{quote}\small\raggedright
\texttt{n\_candidates}, \texttt{n\_history},
\texttt{history\_top1}, \texttt{history\_margin},
\texttt{history\_mean}, \texttt{history\_sd},
\texttt{backbone\_top1}, \texttt{backbone\_margin},
\texttt{sparse}, \texttt{multi\_intent}.
\end{quote}
History count is capped at 30. History scores are candidate cosine
similarities to the normalized mean of history item embeddings; their
maximum, top-two gap, mean, and population standard deviation form four
features. Backbone scores are standardized within each candidate slate
using its mean and population standard deviation, with $10^{-12}$ added
to the denominator; only the maximum and top-two gap enter the router.
A single-candidate gap is zero. Sparse and multi-intent flags come from the
frozen, source-specific history construction. Amazon's flags use history length
at most four and at least three title-derived subcategories, respectively; these
are also the thresholds used by the original Amazon analysis.
Candidate count uses the slate length, not its labels.
The final StandardScaler and ridge model are fitted only on the 1,600
router-training users, with regularization 1 and no hyperparameter search.
This input boundary is conditional on the offline slates described above.

\subsection{Frozen rules}
All four data-dependent rules use the same audit users. Mean-only approves
when mean reward is positive and its two-sided test has $p<.05$, otherwise
deferring. HTE-only requires both permutation statistics to be available;
it approves if prediction $R^2$ or the GATES-style grouped contrast has
$p<.025$, otherwise deferring. Each permuted HTE statistic refits and credits
the permuted rewards consistently. Missing required evidence is insufficient.
The HTE audit uses five fixed folds and 999 shared user permutations; each
permutation refits all five fold-local scalers and ridge models and is credited
against the same permuted rewards. The small CV pilot uses the same five folds.

The frozen RAISE-v1 rule is a conservative instantiation of the broader audit:
it approves when both component rules
approve, defers when both defer, and records insufficient evidence on
disagreement or missing evidence. The small cross-validated ridge pilot
approves when its audit gain interval lies above $\epsilon=.005$, defers
when its upper bound is at most $\epsilon$, and otherwise records insufficient
evidence. Always-approve and always-defer provide constant references.
Table~\ref{tab:prospectivedecisions} retains every prediction. In particular,
the CV pilot's Amazon GRU decision is an abstention, not an explicit rejection.

\begin{table}[h]
\caption{All prospective decisions, frozen before final policy outcomes.
Insufficient denotes an abstention. The combined rule is the tested v1
agreement rule, distinguished from the measurement protocol and the router.}
\label{tab:prospectivedecisions}
\centering\small
\setlength{\tabcolsep}{3pt}
\begin{tabular}{@{}lllll@{}}
\toprule
Audit rule & MIND GRU & MIND SASRec & Amazon GRU & Amazon SASRec \\
\midrule
Mean-only & Defer & Approve & Approve & Approve \\
HTE-only & Approve & Defer & Approve & Approve \\
CV pilot & Insufficient & Insufficient & Insufficient & Approve \\
RAISE-v1 & Insufficient & Insufficient & Approve & Approve \\
Always approve & Approve & Approve & Approve & Approve \\
Always defer & Defer & Defer & Defer & Defer \\
\bottomrule
\end{tabular}

\end{table}

\subsection{Freeze chronology and endpoint reconstruction}
Archived event records place the audit freeze before router fitting, selection
locking, and evaluation scoring.
Table~\ref{tab:prospectivechronology} gives the central boundaries. These
timestamps document artifact chronology; cached replay verifies the numerical
results but does not independently establish the complete history of analyst
access to outcomes.

\begin{table}[h]
\caption{Prospective freeze boundaries in UTC, rounded to seconds. The
archived records retain finer precision and intermediate author confirmation,
router-model, final-outcome, and archive seals.}
\label{tab:prospectivechronology}
\centering\small
\begin{tabular}{@{}lll@{}}
\toprule
Event & MIND & Amazon BPC \\
\midrule
Audit predictions frozen & Sep 15, 07:22:18 & Sep 16, 06:56:24 \\
Evaluation selections locked & Sep 16, 01:32:48 & Sep 16, 17:27:49 \\
Evaluation rewards scored/frozen & Sep 16, 01:33:28 & Sep 16, 17:29:51 \\
\bottomrule
\end{tabular}
\end{table}

For a locked mask $s_i$ and paired reward $\Delta_i$, the reported estimate is
$\widehat G_{.30}=1600^{-1}\sum_i(s_i-.30)\Delta_i$.
All four masks have exactly 480 selected users. The 10,000 shared user
bootstrap draws resample these centered contributions, conditioning on the
fixed models and masks. Resampled selection rates may vary; the frozen
reference rate remains $.30$. The procedure does not refit the router or
reselect a new top-budget cohort inside each bootstrap draw. Its intervals
are consequently fixed-policy sampling summaries, not refit-aware or
distribution-free guarantees.

Numerical reconstruction uses cached paired rewards, selections, seed
20260917, and the shared index manifest. All four gains, interval endpoints,
and bootstrap replicate arrays match the delivered exports. The index-matrix
SHA-256 is recorded in \path{data/followup_20260916.json}. No new generation,
reward-model training, or outcome-dependent reselection enters this replay.

\begin{table}[h]
\caption{Complete prospective ridge outcomes in absolute NDCG@10 units.
The target $\epsilon=.005$, primary designation, and classifications are
unchanged from the frozen evaluation. Intervals are marginal and conditional
on the fitted policy, not a four-endpoint family-wise claim.}
\label{tab:prospectiveresults}
\centering\small
\setlength{\tabcolsep}{3pt}
\begin{tabular}{@{}llrll@{}}
\toprule
Setting & Role & $G_{.30}$ & 95\% interval & At $\epsilon=.005$ \\
\midrule
MIND / GRU & Primary & $+0.005427$ & $[+0.001288,+0.009739]$ & unresolved \\
MIND / SASRec & Secondary & $+0.004966$ & $[+0.000173,+0.009863]$ & unresolved \\
Amazon BPC / GRU & Secondary & $+0.015871$ & $[+0.007619,+0.024372]$ & confirmed-useful \\
Amazon BPC / SASRec & Secondary & $+0.017183$ & $[+0.009373,+0.025278]$ & confirmed-useful \\
\bottomrule
\end{tabular}

\end{table}

In Table~\ref{tab:prospectiveresults}, both MIND intervals exceed zero but
cross the target. They support positive
selection under the reported fixed-policy analysis and leave target
attainment unresolved. Both Amazon intervals exceed the target. Because
RAISE-v1, mean-only, HTE-only, and always-approve all approve both Amazon
conditions, these outcomes do not distinguish their predictive value.
There are two correlated confirmed-useful secondary conditions and zero
resolved below-target conditions. The latter denominator is unavailable
for evaluating false-approval discrimination. Unresolved MIND outcomes
are neither successes nor errors for the target-based audit comparison.
These literal denominators prevent a four-condition pilot from being
interpreted as framework-wide calibration.

\section{Mean-Power Planning and Sensitivity}
\label{app:power_sens}
Let $\sigma_\Delta>0$. Under a normal approximation, the standardized sample
mean is approximately normal with noncentrality
$\sqrt{N}\mu_\Delta/\sigma_\Delta$. A two-sided level-$\alpha$ test rejects
when $|Z|>z_{1-\alpha/2}$. For a positive alternative, the usual planning
approximation neglects the small opposite tail and solves
$\sqrt{N}\rho\approx z_{1-\alpha/2}+z_{1-\beta}$ for target power
$1-\beta$. With two-sided $\alpha=.05$ and 80\% planned power this yields the
mean-effect planning reference quoted in Section~\ref{sec:audit},
\begin{equation}
\rhostar(N)=\frac{z_{1-\alpha/2}+z_{1-\beta}}{\sqrt{N}}
\approx\frac{2.8}{\sqrt{N}},\qquad
N_{\min}\approx\left(\frac{2.8}{\rewsnr}\right)^2.
\label{eq:quickdiag}
\end{equation}
Estimated effects, variance,
and dependence introduce uncertainty into the plug-in $N_{\min}$.
Reaching $N_{\min}$ is neither necessary nor sufficient for useful routing:
it concerns the mean alone, and Section~\ref{sec:theory} shows that the mean
does not determine selection gain.

\begin{table}[h]
\caption{Sensitivity of $\rho/\rho^\star$ to the planning convention. Original
cohorts use legacy mean-SNRs $0.048$, $0.014$, and $0.138$ with
$N=1263$, $650$, and $498$. Fresh GRU/SASRec use $0.1030/0.0674$ at
$N=4954$. These screen mean detectability only; the independent heterogeneity
and held-out policy analyses provide the selective-acquisition evidence.}
\label{tab:power_sens}
\centering\small
\setlength{\tabcolsep}{4pt}
\begin{tabular}{lrrrrrr}
\toprule
$(\alpha,\mathrm{power})$ & Constant & MIND old & Amazon & REES46 & Fresh GRU & Fresh SAS \\
\midrule
$(0.10,0.80)$ & 2.486 & 0.69 & 0.14 & 1.24 & 2.92 & 1.91 \\
$(0.05,0.80)$ & 2.802 & 0.61 & 0.13 & 1.10 & 2.59 & 1.69 \\
$(0.10,0.90)$ & 2.926 & 0.58 & 0.12 & 1.05 & 2.48 & 1.62 \\
$(0.05,0.90)$ & 3.242 & 0.53 & 0.11 & 0.95 & 2.24 & 1.46 \\
$(0.01,0.80)$ & 3.417 & 0.50 & 0.10 & 0.90 & 2.12 & 1.39 \\
\bottomrule
\end{tabular}

\end{table}

In Table~\ref{tab:power_sens}, original MIND and Amazon-Beauty remain below
all listed references. REES46 is
near the boundary and changes classification as the convention changes; its
legacy estimated mean is negative. That sign is relevant to acquire-all,
without ruling out a separately validated positive subgroup. The full fresh
MIND means exceed each displayed planning reference. This stability does not
change their policy intervals or establish routing through the mean screen.

At $N=500$, fresh GRU has $\rho=0.0917$ and
$\rho/\rho^\star=0.732$, while its mean-lift interval is positive. There is
no inconsistency: the normal rejection reference is approximately
$1.96/\sqrt{N}$, whereas $2.8/\sqrt{N}$ plans for 80\% rejection
probability over repeated samples. Clearing the first in one sample need not
clear the second.

\section{Cost Interpretation of the Fresh-Cohort Gains}
\label{app:costfrontier}
For fixed $0<b<1$, write $V_b=\E[\pi_b\Delta]=b\mu_\Delta+G_b$.
Under a uniform reward-equivalent call cost $\lambda\ge0$ and no routing overhead,
\[
U(\pi_b)=V_b-b\lambda,\qquad
U(\mathrm{all})=\mu_\Delta-\lambda,\qquad
U(\mathrm{skip})=0.
\]
The fitted policy beats both all and skip when
\begin{equation}
\frac{\mu_\Delta-V_b}{1-b}<\lambda<\frac{V_b}{b}.
\label{eq:costinterval}
\end{equation}
The interval in Eq.~\ref{eq:costinterval} compares fixed policies under an
illustrative uniform cost. Appendix~\ref{app:cost} reports the original
generation latency that motivates treating the call as costly.
The follow-up recomputes value at the exact observed acquisition fraction,
$q=1486/4954\approx0.29996$, rather than rounding it to $0.30$
(Table~\ref{tab:costpoints}).

\begin{table}[h]
\caption{Fresh-MIND ridge incremental value retained at the actual acquisition rate.
Intervals jointly resample selected and acquire-all rewards over 10,000 user
draws, conditional on fitted policies. Retained value is $V_q/\mu_\Delta$;
it includes the $q\mu_\Delta$ random-acquisition component and is not a
fraction of total ranking accuracy or a measured monetary saving.}
\label{tab:costpoints}
\centering\small
\begin{tabular}{@{}lrrrrl@{}}
\toprule
Ridge & Actual $q$ & $G_q$ & $V_q$ & $V_q/\mu_\Delta$ & Joint 95\% interval \\
\midrule
GRU & $0.299960$ & $0.00709$ & $0.01216$ & 71.86\% & $[58.30\%,89.22\%]$ \\
SASRec & $0.299960$ & $0.00517$ & $0.00919$ & 68.51\% & $[48.59\%,99.12\%]$ \\
\bottomrule
\end{tabular}

\end{table}

The cached budget study retains 48 policy/budget cells and 384 uniform-cost
evaluations, including 100 negative utilities. Budgets are 10\%, 20\%, 30\%,
50\%, 75\%, and 100\%; costs range from 0 to 0.05 in reward-equivalent
units. At 100\% acquisition, selection gain is exactly zero. These are offline
value calculations with zero assumed routing overhead, not a learned,
priced deployment frontier. Nonuniform token costs, latency, overhead, and
failure fallback require additional measurements.

\paragraph{Prospective generation and audit accounting.}
MIND records 4,302 requests (4,100 initial and 202 retries), 5,755,211 known
tokens, and 123 requests without token usage. Amazon records 4,149 requests
(4,100 initial and 49 retries), 5,101,068 known tokens, and two requests
without usage. Shared generation is counted once across backbones and audit
methods; reasoning tokens already included in completion usage are not added
again. Missing usage makes known totals lower bounds. The standalone full
audit takes approximately 5.59 seconds on MIND and 6.01 seconds on Amazon,
versus roughly 0.077 seconds for the small CV pilot and 0.0006 seconds for
the mean-only rule in the recorded environment. These timings exclude
generation, reward-model training, and production serving. They do not
establish end-to-end latency or dollar savings.

\section{Structured Hypothesis Embeddings and Average-Use Evidence}
\label{app:she}\label{app:embed}
\begin{figure}[h]
\centering
\includegraphics[width=\textwidth]{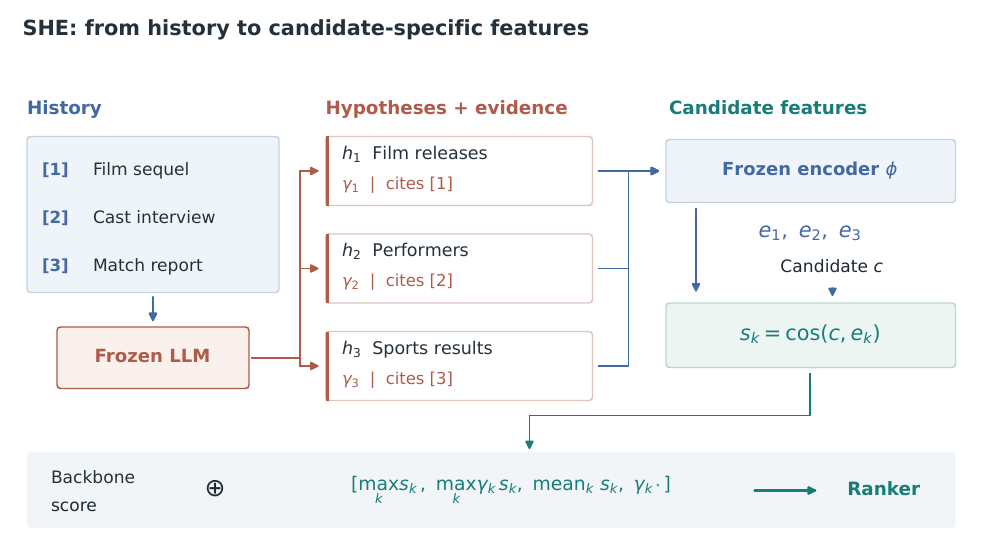}
\caption{\textbf{SHE maps a history to four candidate-specific ranking
features.} A frozen LLM generates three hypotheses with confidence scores and
evidence indices.
Constructed history and hypothesis examples illustrate the format;
confidence scores are symbolic. Frozen embeddings yield
candidate--facet matches, whose four-coordinate block is concatenated with
the backbone score for ranking; the original Amazon-Beauty analysis uses the
legacy variant defined in this appendix. The before-call acquisition decision is
shown separately in Figure~\ref{fig:workflow}.}
\label{fig:pipeline}
\end{figure}

SHE provides a concrete observation whose content can be inspected separately
from its acquisition value. Figure~\ref{fig:pipeline} shows how a history
becomes ranking features. The original analyses below examine citation
alignment, confidence calibration, and average predictive gains across
backbones. Their cohorts are distinct from the fresh-MIND routing study.

\paragraph{Embeddings and feature block.}
The original content-based MIND and REES46 analyses use
\texttt{text-embedding-ada-002}, $d=1536$. Amazon-Beauty and the original
ordered-backbone MIND study use a local 256-dimensional TF-IDF plus
TruncatedSVD (LSA) representation. Fresh MIND also uses a prior-cohort-fitted LSA
representation, frozen before fresh scoring; its configuration and checkpoints
are preserved internally but are not currently public. Hypotheses and items are embedded in the same
space within each analysis, with $\ell_2$ normalization. Cosines are not
compared across different encoders as if they shared a scale.

The four-coordinate SHE block, whose confidence-weighted coordinate
is Eq.~\ref{eq:she}, is
\[
\left[\max_k\cos(\vc,\ve_k),\quad
\max_k\gamma_k\cos(\vc,\ve_k),\quad
\frac1K\sum_k\cos(\vc,\ve_k),\quad
\gamma_{k^\star}\right],
\]
where $k^\star$ maximizes the unweighted facet match. The MIND analyses and
the prospective Amazon pipeline use this block. The original Amazon-Beauty
analysis used a legacy variant,
\[
\left[\max_k\cos(\vc,\ve_k),\quad
\gamma_{k^\star},\quad
\frac1K\sum_k\cos(\vc,\ve_k),\quad
\frac{\sum_k\gamma_k\cos(\vc,\ve_k)}{\sum_k\gamma_k}\right],
\]
consisting of maximum cosine, confidence of the maximum-cosine hypothesis, mean
cosine, and confidence-weighted mean cosine; the last coordinate is zero when all
confidences are zero, and a user without Scheme-B hypotheses receives an
all-zero block. Its reported results retain that original implementation. Each
block is concatenated with the base score. The original MIND late-fusion head is an
$\ell_2$-regularized, class-balanced logistic ranker with $C=1$;
standardization uses training-fold statistics. The original REES46 study
uses a logistic/ridge probe and PCA reduction for its low-$N$ setting.
Downstream gradients do not update the LLM or text encoder.

\subsection{Semantic checks}
\label{sec:agent}\label{app:agent}
On original MIND, hypothesis embeddings are closer to cited than non-cited
evidence by $0.0705$ cosine points, with 95\% CI $[0.068,0.073]$.
The corresponding alignment is approximately zero on REES46, whose windows
are 87.9\% single-category with thin text. Pairwise hypothesis
distinctiveness $1-\overline{\cos}$ is $0.204$ on MIND and $0.104$ on
REES46. These embedding-space checks make citation support inspectable; they
are not human-verified semantic truth labels.

\begin{figure}[h]
\centering
\includegraphics[width=\textwidth]{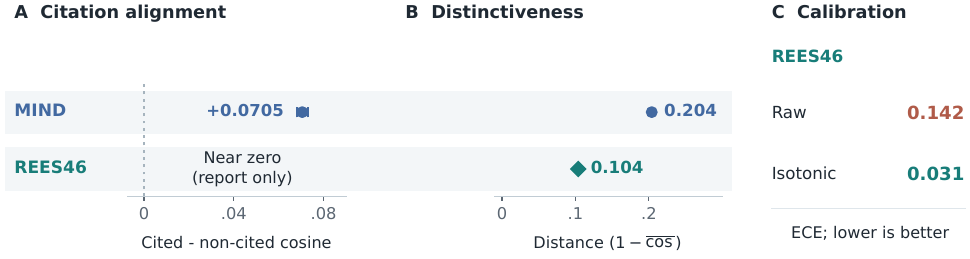}
\caption{\textbf{Original-cohort semantic checks inspect signal content, not
selective value.}
\textbf{A:} MIND citation alignment: mean cited-minus-noncited cosine difference
$0.0705$, with 95\% CI $[0.068,0.073]$; REES46 is reported only as near zero.
\textbf{B:} Mean inter-hypothesis distinctiveness (intervals unavailable).
\textbf{C:} REES46 cited-alignment ECE before and after five-fold cross-fit
isotonic calibration; only aggregate values are available, without bin coordinates or
intervals.}
\label{fig:agent}
\end{figure}

On REES46, five-fold cross-fit isotonic calibration reduces ECE from
$0.142$ to $0.031$ for the cited-evidence-alignment target; the raw Brier
score is $0.166$. Calibration here concerns that target, not the probability
that acquisition improves ranking. Scheme A and Scheme B use different
generation settings, so their comparison is descriptive rather than a
structure-only ablation. Figure~\ref{fig:agent} collects these three semantic
checks.

\subsection{Backbone-conditional average value}
\label{sec:backbone}\label{app:robust}\label{app:amazon}
\begin{figure}[h]
\centering
\includegraphics[width=.84\textwidth]{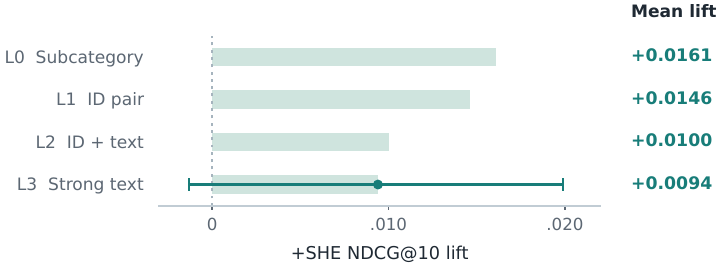}
\caption{\textbf{SHE's average-use point estimates decrease as the original-MIND
baseline becomes richer.} Bars show mean SHE gains across four nested feature sets. Only
L3 has a numerical 95\% CI in the retained summaries; the other bars show point
estimates without intervals. This cohort and reward pipeline differ from fresh
MIND and do not establish selective value.}
\label{fig:ladder}
\end{figure}

Across the original MIND feature ladder, SHE's added mean NDCG@10 lift
decreases from $0.0161$ at L0 to $0.0146$, $0.0100$, and $0.0094$ as
the baseline becomes richer. At L3 its interval is
$[-0.0013,+0.0199]$ (Figure~\ref{fig:ladder}). This is an average-use
redundancy pattern within that analysis, not a routing threshold.

\begin{figure}[!htbp]
\centering
\includegraphics[width=.92\textwidth]{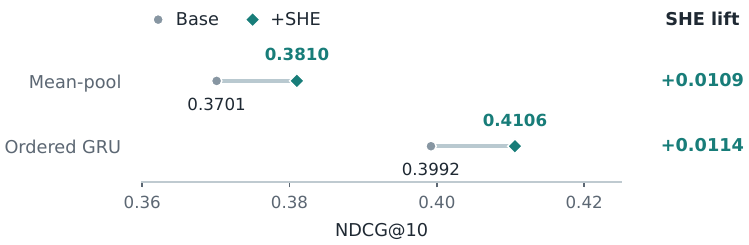}
\caption{\textbf{Positive SHE point estimates do not establish a
backbone-ordering interaction.} In the original-MIND LSA study, connected
points show each backbone before and after adding SHE. The mean-pool minus GRU
lift difference is $-0.0005$ ($p=0.919$); the interaction remains unresolved.}
\label{fig:backbone}
\end{figure}

The ordered/unordered $2\times2$ experiment on 1,411 LSA MIND windows
asks a more specific question. The base mean-pool and GRU scores are
$0.3701$ and $0.3992$ NDCG@10. Adding SHE yields $+0.0109$
$[-0.0046,+0.0277]$ over mean-pool and $+0.0114$
$[+0.0030,+0.0209]$ over the ordered GRU. The interaction is
$-0.0005$ $[-0.0164,+0.0150]$, $p=0.919$ (Figure~\ref{fig:backbone}). Thus the
data do not establish a different SHE effect between these two backbones even
though one individual interval is positive and the other crosses zero.

Original SASRec robustness shows $+0.0179$ over its ordered backbone.
On Amazon-Beauty (median history length five), the evaluated GRU score
$0.4104$ is below mean-pool's $0.4777$; ordering alone need not make
this fitted backbone stronger. These observations motivate distinguishing
sequence availability, backbone strength, and effect size. They are
exploratory signal studies rather than evidence that sequential versus
nonsequential data universally determine the benefit of SHE.

\paragraph{Exploratory coarse gate.}
The older sparse/multi-intent gate acquires on 86.4\% of original MIND
windows. Its contrast with acquire-all is approximately $+0.0001$ with
95\% CI $[-0.0038,+0.0042]$. This interval permits gains and losses
and does not establish equivalence. The gate was derived from the same
data and needs independent validation. Fresh MIND shows that a coarse gate is not
the only plausible deployment policy: per-instance routers can also recover
value in an appropriate setting.

\section{Raw Input Examples}
\label{app:rawinputs}
\begin{figure}[h]
\centering
\includegraphics[width=\textwidth]{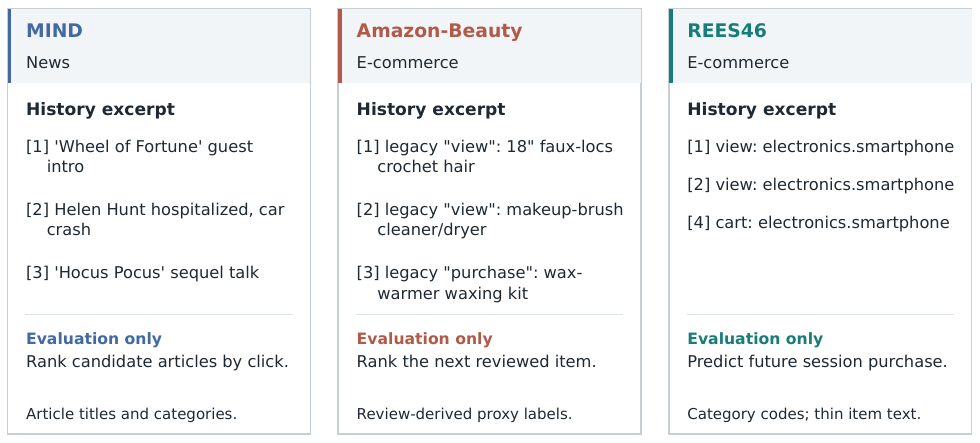}
\caption{\textbf{Original-dataset histories differ in semantic richness.}
Abbreviated excerpts illustrate title-rich MIND and Amazon-Beauty histories
and a category-code-dominated REES46 window. Amazon's displayed
\texttt{view}/\texttt{purchase} labels are legacy review-derived proxies, not
logged actions; its target is the next reviewed item. The lower row describes
the evaluation task, separately from the pre-target history supplied to the
LLM. These qualitative excerpts provide context rather than performance evidence.}
\label{fig:rawinputs}
\end{figure}

Figure~\ref{fig:rawinputs} shows the input histories supplied to the frozen LLM
in each original dataset; Appendix~\ref{app:prompts} reproduces the verbatim
generation prompts.

\section{SHE Prompts and Generation Settings}
\label{app:prompts}

The original study uses fixed zero-shot prompts to turn histories into
structured intent hypotheses. HypoGeneAgent~\citep{hypogeneagent2025} uses a
related hypothesis, confidence, and evidence format for gene-set annotation;
SHE applies that format to behavioral histories. The prompts request
confidence scores, whose empirical calibration is evaluated separately.
The fresh-MIND generation manifest is preserved internally but is not currently
public.

\paragraph{Decoding and model configuration.}
Scheme~A returns one blended-intent summary; Scheme~B returns the ranked,
evidence-indexed hypotheses used by SHE. Table~\ref{tab:agent_cfg} lists the
reported original settings. Both use the same API transport, without
fine-tuning, retrieval, or tool use. Their different model versions and
reasoning effort make the comparison a joint configuration contrast.

\begin{table}[h]
\caption{Generation settings for the two hypothesis schemes. Scheme~B
is the source of the SHE feature block (App.~\ref{app:embed}); Scheme~A is a
descriptive summary baseline, not a matched structure-only ablation.}
\label{tab:agent_cfg}
\centering
\small
\begin{tabular}{lll}
\toprule
 & \textbf{Scheme A (summary)} & \textbf{Scheme B (SHE, Top-3)} \\
\midrule
Model            & GPT-5.4                       & GPT-5.5 \\
Reasoning effort & low                           & high \\
Output           & \verb|{"summary": ...}|       & \verb|{"hypotheses": [...]}| \\
\# hypotheses    & 1 (free sentence)             & exactly 3, rank-ordered \\
Per-item fields  & ---                           & hypothesis, confidence, evidence\_indices \\
Fine-tuned?      & no (zero-shot)                & no (zero-shot) \\
\bottomrule
\end{tabular}
\end{table}

\paragraph{Input serialization.}
Numbered pre-target events form the history input. MIND uses clicked article
titles and categories. REES46 supplies logged actions and available category or
item descriptions. Amazon instead supplies review-derived \texttt{view} and
\texttt{purchase} proxy tokens, defined in Appendix~\ref{app:data}. The step index
\texttt{[i]} anchors each returned \texttt{evidence\_indices} entry. The excerpts
in Appendix~\ref{app:rawinputs} illustrate the available content; evaluation
targets are kept separate from the history prompt.

\paragraph{Output schema (Scheme~B).}
A single JSON object contains exactly three rank-ordered hypotheses. Each
has a \texttt{hypothesis} statement, a model-reported \texttt{confidence} in
$[0,1]$, and \texttt{evidence\_indices} pointing to supporting history events.
The word \emph{calibrated} in the prompt specifies a request, not an empirical
guarantee. The fixed hypothesis slots produce a consistent input to the
four scalar SHE features in Appendix~\ref{app:embed}.

\paragraph{Boundary rules.}
The domain prompts specify weak-signal fallbacks, evidence requirements, and
restrictions on unsupported demographic inferences. For weak histories,
scores must remain below $0.5$ and the third hypothesis uses a fixed browsing
fallback. The e-commerce prompt requires an add-to-cart trigger for the top
scores to exceed $0.7$; the news prompt asks for distinct facets when several
topics appear. These are generation instructions. Their joint output is
assessed by the semantic checks in Appendix~\ref{sec:agent}; separate rule
ablations would be needed to attribute the results to individual instructions.
For Amazon, the frozen input has no add-to-cart actions because it is derived
from reviews, not shopping-session logs. Nevertheless, 22 of 650 cached outputs
place a top-two confidence above $0.7$. The mismatch between the verbatim
shopping-window prompt and these review-derived proxies limits semantic
interpretation; the original prompt and cache are retained for provenance.

\paragraph{Verbatim prompts.}
The complete system prompts follow. The serialized window (above) is appended as
the user turn.

\vspace{2pt}
\noindent\textbf{News --- Scheme A (single summary).}
\begin{lstlisting}[style=prompt]
You are a news-recommendation and reading-interest expert. Below is a user's recent [reading history] (news articles they clicked, oldest to newest). In ONE free-text sentence of at most 30 words, give a single sharp summary of the user's current core reading interest. Output STRICTLY a JSON object: {"summary": "<your summary>"} -- no markdown, no extra explanation.
\end{lstlisting}

\noindent\textbf{News --- Scheme B (SHE, Top-3 ranked hypotheses).}
\begin{lstlisting}[style=prompt]
You are a top expert in news-recommendation and reader psychology. Below is a user's recent [reading history] (clicked news articles, oldest to newest). Based strictly on the reading evidence, infer the user's Top-3 ranked [Reading-Interest Hypotheses] -- the distinct themes/topics the user is most likely to want to read next.
Output STRICTLY a single JSON object: {"hypotheses": [ ... ]} whose array holds EXACTLY 3 elements, ordered by descending strength. No markdown (no ```json), no extra explanation. Each element must contain these fixed keys:
1. "hypothesis": a specific reading-interest statement with reasoning about the theme.
2. "confidence": a calibrated probability score between 0.0 and 1.0.
3. "evidence_indices": an integer array of the input step indices (the [n] at the start of each input line) that support this hypothesis.
[Hard boundary rules]
- Weak-signal fallback: if the history is very short or all one narrow topic, every confidence must be below 0.5, and the 3rd hypothesis must be exactly "Casual browsing / no strong topical interest".
- Multi-interest capture: when the history spans several distinct topics, the three hypotheses MUST cover genuinely different interest facets (do not restate one topic).
- Bias elimination: never fabricate the user's absolute gender, age, or region.
\end{lstlisting}

\noindent\textbf{E-commerce --- Scheme A (single summary).}
\begin{lstlisting}[style=prompt]
You are an e-commerce consumer-behavior expert. Below is a user's [behavior sequence] within a single shopping window. In ONE free-text sentence of at most 30 words, give a single sharp summary of the user's current core shopping intent. Output STRICTLY a JSON object: {"summary": "<your summary>"} -- no markdown, no extra explanation.
\end{lstlisting}

\noindent\textbf{E-commerce --- Scheme B (SHE, Top-3 ranked hypotheses).}
\begin{lstlisting}[style=prompt]
You are a top expert in e-commerce consumer behavior and user psychology. Below is a user's [behavior sequence] within a single shopping window. Based strictly on the concrete behavioral evidence, infer the user's Top-3 most urgent [Ranked Intent Hypotheses] (immediate purchase motivations).
Output STRICTLY a single JSON object: {"hypotheses": [ ... ]} whose array holds EXACTLY 3 elements, ordered by descending urgency. No markdown (no ```json), no extra explanation. Each element must contain these fixed keys:
1. "hypothesis": a specific intent-hypothesis statement with behavioral-motivation reasoning.
2. "confidence": a calibrated probability score between 0.0 and 1.0.
3. "evidence_indices": an integer array of the input step indices (the [n] at the start of each input line) that support this hypothesis.
[Hard boundary rules]
- Weak-signal fallback: if the input is ALL views (view) with NO add-to-cart, every confidence must be below 0.5, and the 3rd hypothesis must be exactly "Just browsing / no clear purchase goal".
- Strong-signal trigger: only when an add-to-cart action is present may the top two hypotheses exceed 0.7 confidence.
- Bias elimination: never fabricate the user's absolute gender, age, or region from category names.
\end{lstlisting}

\section{Original-Study Window Construction and Settings}
\label{app:data}

This section describes window construction for the original signal and
backbone studies; the fresh-MIND protocol is in Appendix~\ref{app:f4}.
All three datasets are public. A \emph{window} pairs a pre-target user history
with an evaluation target, and hypothesis generation uses the history.
Table~\ref{tab:windows} gives the exact construction. A ``window'' is a
point-in-time slice: the history is logged behavior for MIND and REES46 but a
chronological review history for Amazon. The target is a future click (MIND
candidate slate), a future purchase (REES46 session), or the next reviewed item
(Amazon-Beauty). Cold-start (\emph{sparse}) and
\emph{multi-intent} slices are defined by fixed thresholds, not tuned.

\begin{table}[h]
\caption{Window construction per dataset. MIND is built with a cohort of 1{,}600
impressions at stride 5 (keep 1 of every 5) and a 40-candidate slate cap;
hypotheses were generated for 1{,}557 windows (1{,}549 valid for both schemes).
Amazon-Beauty: 650 windows (222 sparse; 420 multi-intent). REES46: 498 session windows. History
length is the number of pre-target actions; the \emph{sparse}/\emph{multi} slices
are the cold-start and multi-intent regimes analyzed throughout.}
\label{tab:windows}
\centering
\small
\begin{tabular}{llll}
\toprule
 & \textbf{MIND (news)} & \textbf{Amazon-Beauty} & \textbf{REES46} \\
\midrule
Domain              & news reading       & e-commerce         & e-commerce \\
Target              & candidate click    & next reviewed item & session purchase \\
Windows generated   & 1{,}557            & 650                & 498 \\
History length      & median 19, mean 23.6 & median 5         & median $\sim$9 \\
Slate size (cap)    & 40 candidates      & 20 candidates      & --- (binary) \\
\emph{sparse} rule  & $n_{\text{hist}}\le 5$ & $n_{\text{hist}}\le 4$ & $n_{\text{hist}}\le 5$ \\
\emph{multi} rule   & $\ge 8$ distinct cats & $\ge 3$ distinct subcats & (mostly single-cat) \\
Content richness    & high (multi-topic) & high (titles)      & low (87.9\% single-cat) \\
\bottomrule
\end{tabular}
\end{table}

\paragraph{Amazon review-derived action proxies.}
Amazon-Beauty does not contain logged view, cart, and purchase events. The
frozen preprocessing maps a review to \texttt{purchase} only when
\texttt{verified\_purchase} is true and rating is at least four; every other
review is mapped to \texttt{view}. Across the 650 histories this yields 1,024
\texttt{purchase} and 2,688 \texttt{view} tokens; 323 verified low-rating
reviews are therefore labeled \texttt{view}. No \texttt{cart} token occurs.
The median history spans 409 days, so the serialized sequence is not literally
a single shopping session. These tokens are legacy behavioral proxies, and the
task is next-reviewed-item ranking, not purchase-event prediction. The frozen
cache and results retain this mapping rather than retroactively reserializing
the input.

\paragraph{Per-analysis $N$.}
The number of \emph{rerankable} windows in a given result can be smaller than the
generated count, and eligibility is analysis-specific. Windows with an empty valid
slate are dropped, and the MIND analyses exclude windows without valid hypotheses.
The original Amazon feature pipeline instead retained four users without Scheme-B
hypotheses with a zero-valued SHE block, giving $N{=}650$; the later prospective
Amazon historical fitting pipeline excluded those users and used 646 eligible
All\_Beauty users. Prospective generation failures follow their own
complete-generation eligibility rules (Appendix~\ref{app:prospective}). We
therefore report the exact $N$ with each result
(e.g.\ MIND $N{=}1263$ for the ada-002 late-fusion study, $N{=}1411$ for the
backbone-conditional LSA study; Amazon $N{=}650$; REES46 $N{=}498$). No window is
dropped for having a favorable or unfavorable reward under the reported
eligibility rules. Verifying the exclusions requires the original row manifests.

\paragraph{Statistical unit.}
Unless otherwise stated, $N$ denotes evaluation windows/impressions and all
reported bootstrap intervals resample at that level. The reward-SNR
$N_{\min}$ values should therefore be read as point-estimate diagnostics under
the window-level independence approximation. If multiple windows share a user,
overlap in history, or share unusually many candidate items, the effective sample
size may be smaller. A user/session-clustered uncertainty audit is the right
confirmatory analysis for deployment; we report the window-level diagnostic and
treat clustered effective-$N$ as follow-up uncertainty analysis.

\section{Hyperparameters}
\label{app:hparams}

Table~\ref{tab:hparams} retains the reported original-study hyperparameters.
These settings should not be taken as a verified configuration manifest for fresh MIND,
whose backbones and heads are fitted on a prior cohort and then frozen.

\begin{table}[h]
\caption{Original-study downstream hyperparameters. The ranker, backbones, and LSA space are
fixed a priori; the GRU/SASRec inputs are the frozen LSA item vectors, so only the
sequence-combination parameters are learned. All CIs use the same impression-level
bootstrap.}
\label{tab:hparams}
\centering
\small
\begin{tabular}{lll}
\toprule
\textbf{Component} & \textbf{Setting} & \textbf{Value} \\
\midrule
Late-fusion ranker   & model              & logistic regression (pointwise) \\
                     & regularization $C$ & $1.0$ \\
                     & max iterations     & $2000$ \\
                     & class weighting    & balanced \\
                     & feature scaling    & StandardScaler, fit on train fold only \\
Evaluation           & cross-fit          & GroupKFold(5), grouped by impression \\
                     & prediction         & out-of-fold click probability $\to$ rank \\
                     & metrics            & NDCG@10, Recall@10, MRR (per impression) \\
                     & confidence interval& bootstrap 95\%, 2000 resamples by impression \\
Item text space (LSA)& encoder            & TF-IDF $+$ TruncatedSVD \\
                     & dimensions         & $256$ (random\_state $=0$) \\
GRU backbone         & hidden size        & $64$ \\
                     & optimizer          & Adam, lr $10^{-2}$, weight decay $10^{-4}$ \\
                     & epochs             & $15$, cross-fit (OOF) \\
SASRec backbone      & attention          & $2$ heads, $1$ layer, max length $50$ \\
                     & hidden size        & $64$ \\
                     & optimizer          & Adam, lr $10^{-3}$, weight decay $10^{-4}$ \\
                     & epochs             & $15$, cross-fit (OOF) \\
Robustness runs      & seeds              & $\{0,1,2,3,4\}$ \\
\bottomrule
\end{tabular}
\end{table}

\paragraph{Fresh and prospective configurations.}
The preserved fresh-MIND run fits an at-most-20,000-feature unigram/bigram TF--IDF
encoder and 256-dimensional TruncatedSVD ($\texttt{random\_state}=0$) on text
from all 1,557 prior windows. Backbones and ranking heads use the 1,549 eligible
windows after excluding eight without valid Scheme-B hypotheses. Both backbones
train for 15 epochs: GRU uses 64 hidden units and Adam with learning rate
$10^{-2}$ and weight decay $10^{-4}$, while SASRec uses a 256-dimensional
attention representation with a 64-unit feed-forward layer, two heads, one layer,
maximum length 50, Adam with learning rate $10^{-3}$, and the same weight decay.
For each backbone, separate base and SHE-augmented pointwise logistic heads
($C=1$, balanced classes, 2,000 maximum iterations) are fitted on the same 1,549
windows and frozen for all 4,954 fresh users. Seed 202609 identifies
this run. The prospective MIND reward pipeline uses the same LSA dimension,
backbone dimensions, epoch count, and prior-cohort training hash; Amazon uses
the same dimensions and training schedule in its separately frozen All\_Beauty
pipeline. Router configuration,
five-fold audits, 999 permutations, tie handling, and checkpoint-to-cohort
boundaries are specified in Appendix~\ref{app:prospectiveconfig}. The source
snapshots, manifests, and checkpoints are preserved internally but are not
currently public.

\section{Hypothesis-Generation Cost and Latency}
\label{app:cost}

The original generation runs illustrate why SHE is a costly observation.
Table~\ref{tab:cost} retains their reported latency measurements. They are
historical timings for those generation configurations, not fresh-MIND cost
measurements. Fixed-budget routing compares how effectively the call budget is
used; a complete serving-cost account must additionally include embeddings,
router inference, retries, and other overhead. Appendix~\ref{app:costfrontier}
separates that accounting from reward-equivalent break-even calculations.

\begin{table}[h]
\caption{Reported original-generation latency. Scheme~B (the SHE source) uses
the high-effort setting and has longer mean call latency. The per-window
wall-clock figures are measured end-to-end and include API response time plus
orchestration overhead, serialization, rate-limit/retry backoff, and worker idle
time, so they need not equal the arithmetic sum of the mean per-call latencies.
Amazon-Beauty's 650 windows took 284.8\,min on a single worker; MIND's 1{,}557
windows were generated across 8 parallel workers (\,$\sim$63\,min wall\,). All
downstream analyses use cached hypotheses. Independent reproduction additionally
requires the corresponding code, cached inputs, and configuration artifacts.}
\label{tab:cost}
\centering
\small
\begin{tabular}{llll}
\toprule
 & \textbf{Scheme A} & \textbf{Scheme B} & \textbf{Per window (A$+$B)} \\
\midrule
Model            & GPT-5.4       & GPT-5.5           & --- \\
Reasoning effort & low           & high              & --- \\
Latency / call   & $\sim$2--3\,s & $\sim$6--17\,s    & --- \\
MIND             & ---           & ---               & $\sim$19--23\,s/win \\
Amazon-Beauty    & 3.0\,s (mean) & 8.6\,s (mean)     & 26.3\,s/win \\
\bottomrule
\end{tabular}
\end{table}

The exploratory original-cohort gate is summarized in
Appendix~\ref{sec:backbone}. It remains a candidate coarse policy requiring
independent evaluation, not a conclusion that only design-time gating is
possible. Fresh-MIND selective policy results are reported separately in
Section~\ref{sec:fresh}.

\section{Evidence Provenance and Remaining Checks}
\label{app:provenance}\label{app:reviewer}
The frozen analysis record of the original study round supplies the initial
follow-up evidence. The figure builder,
\path{scripts/build_revision_assets.py}, parses its tables, checks their
internal arithmetic, and produces vector PDFs and TeX tables. The extracted
values, source-table locations, and record checksum are recorded in
\path{data/f1_f4_report_values.json}. This makes the presentation traceable;
source-code and raw-output reproduction remain separate checks.

The completed follow-up delivers ten frozen packages with source snapshots,
cached outputs, split manifests, model artifacts, analysis assets, and checksums.
They document the controlled screen's binary rule, the before-call allowlists, removed post-call
fields, retrospective refits, and prospective audit/policy boundaries.
\path{scripts/build_followup_assets.py} verifies the five package hashes used
for its exports, reads their CSV/JSON/NPZ files, and generates additional
vector figures and TeX tables. The four prospective gains and all 10,000
bootstrap replicates per condition have been reproduced from cached rewards
and locked selections; provenance is recorded in
\path{data/followup_20260916.json}. This numerical replay is narrower than
rerunning LLM generation, reward-model training, or every scientific refit.

The controlled figure uses the six reported corrected cases. Earlier phase
curves and an earlier agentic-router experiment are excluded pending
verification of their result tables and feature state. Original oracle and average-use summaries
retain their original-cohort scope. Later mechanism and prospective experiments
are reported with their own reward construction and sample boundaries.

For the original figures, \texttt{data/original\_figure\_summaries.json}
records source locations for the reported calibration errors and feature-ladder
means. The current plots show these summaries directly. Bin-level calibration
coordinates, sparse-slice estimates, degradation-sweep tables, and most
ladder intervals are unavailable in the retained numerical materials, so
curves and error bars for those quantities are omitted. The L3 interval is
preserved from the original appendix. This replaces hand-entered plot arrays
with traceable summaries, without implying raw-data reproduction.

The completed retained-value analysis supplies joint uncertainty from cached
rewards. No estimate is reconstructed from a rendered image. The historical
sample-size plot retains reported mean intervals and counts; the new fixed-test
plot uses the delivered numerical learning curve. All eight fresh-MIND policies and
all four prospective conditions remain in the record. Formal
permutation and family-wise inference remains open for the reason in
Appendix~\ref{app:inferencegap}; its historical labels are not adopted as
confirmed significance claims. The reconstruction packages, fresh-MIND
generation manifest, and upstream configurations/checkpoints described above
are preserved internally and are not currently public. Code will be released upon acceptance.

\section{Further Connections to Prior Work}
\label{app:related}
These connections guide policy choice, evaluation, and extensions of the audit.

\paragraph{Active information acquisition.}
Active feature acquisition selects measurements given currently observed
information~\citep{aronsson2025activefeature}. EDDI combines a
partial variational autoencoder with expected information gain
\citep{ma2019eddi}. Generative surrogate models provide acquisition rewards
and state information for sequential feature selection~\citep{li2021gsm}.
RAISE can audit such a choice once the observation, downstream predictor,
and cost are specified. Our experiments cover one optional observation block;
multiple adaptive acquisitions would require history-indexed rewards,
costs, and allowed information.

\paragraph{Information acquisition within an agent.}
\label{app:agentinterface}
ReAct interleaves reasoning with actions that obtain external information
\citep{yao2023react}. To audit an acquisition decision within such a system,
an adapter would expose the allowed before-call context, the optional observation,
the downstream decision rule, and the cost model. The gate sees only the
before-call context; the downstream rule receives the observation if acquired.
Audit data must separately supply independently credited outcomes for the
specified acquire/skip comparison. The audit budget includes paid calls
needed to collect this initial evidence.
For a sequential extension, the comparison must specify the decision history
and continuation policy, since an observation can change later actions and
future observation opportunities. Repeated simulator rollouts could estimate
branch returns; logged interactions instead require suitable off-policy
identification and action support. Training and evaluation must separate
trajectories or independent task instances, with uncertainty accounting for
within-trajectory dependence. Cost accounting must include delays and subsequent
calls. This proposed integration interface specifies the information, reward,
and cost records needed for an agent acquisition audit. The empirical studies
instantiate the single-step setting through optional SHE acquisition and ranking.

\paragraph{Deferral, rejection, and incremental benefit.}
Selective classification trades coverage for predictive risk
\citep{geifman2017selective}. Learning to defer models the downstream
decision-maker's performance~\citep{madras2018defer}; consistent surrogate
losses support joint learning of a predictor and a deferral rule
\citep{mozannar2020defer}. Base-predictor uncertainty need not identify where
an additional observation improves the outcome. RAISE measures this incremental
benefit conditional on available information; budget-matched random acquisition
separates selecting well from using the resource more often.

\paragraph{Model pools, cascades, and the information available to a router.}
FrugalGPT learns cascades that can inspect intermediate responses
\citep{chen2024frugalgpt}; RouteLLM learns query-based model selection from
preference data~\citep{ong2024routellm}. Cascade routing provides a unified
optimization view~\citep{dekoninck2025cascade}, while BEST-Route jointly
chooses the model and the number of candidate responses
\citep{ding2025bestroute}. Self-REF learns confidence tokens that inform
downstream routing and rejection~\citep{chuang2025selfref}. An intermediate
response can guide a later escalation, with its generation cost included.
Each gate uses information available before its own call; the feature
allowlist in Protocol~\ref{prot:raise} records this timing boundary for
each decision being evaluated.

\paragraph{Benchmarks, risk constraints, and deployment infrastructure.}
RouterBench supplies model outcomes for reproducible routing comparisons
\citep{hu2024routerbench}. LLMRouterBench broadens the model and task coverage
and reports both simple-baseline competitiveness and remaining oracle gaps
\citep{li2026llmrouterbench}. LLMRouter provides modular routing components
and evaluation across additional settings, including personalized and
multi-turn tasks~\citep{feng2026llmrouter}. These resources could support
wider audits. Conformal LLM
routing calibrates a gate for the probability that the cheap model is wrong
while the expensive reference is correct among cheap-routed queries, under
its exchangeability assumptions~\citep{uddin2026conformal}. This conditional
violation target differs from average selection gain at a fixed budget.
RAISE reports empirical selection gains with setting-specific uncertainty;
distribution-free risk control would require an additional calibration guarantee.

\paragraph{Estimating heterogeneity and evaluating a decision rule.}
Causal forests support heterogeneous-effect estimation and inference
\citep{wager2018causalforest}; meta-learners adapt supervised learners to
treatment-effect estimation~\citep{kunzel2019metalearners}, and the R-learner
separates nuisance estimation from the effect-learning objective
\citep{nie2021quasioracle}. The quality of an effect estimate and the quality
of a resulting action rule remain different objectives
\citep{fernandezloria2022causal}. Empirical welfare maximization and policy
learning optimize treatment allocation~\citep{kitagawa2018welfare,athey2021policy}.
PAPE evaluates improvement over random allocation with the same treatment
rate~\citep{imai2023experimental}; RATE evaluates the benefit concentration
induced by a prioritization score~\citep{yadlowsky2025rate}. GATES-style grouping helps
inspect predictable variation~\citep{chernozhukov2018generic}, while direct
held-out evaluation determines the evidence for an actual acquisition rule.
RAISE uses mean and regression-fit diagnostics to characterize the reward
channel, and measures a policy's selection gain directly on held-out rewards.

\paragraph{Offline ranking contrasts and off-policy evaluation.}
With logged bandit feedback, only the chosen action's outcome is observed.
Doubly robust policy evaluation combines reward and propensity models
\citep{dudik2011doublyrobust}; counterfactual risk minimization accounts for
the variance of propensity-weighted objectives when learning a policy
\citep{swaminathan2015crm}. Our real-data studies instead score both ranking
pipelines against retained relevance labels on the same audited examples.
The resulting contrast measures incremental offline ranking performance.
Estimating effects on production outcomes from logged interactions would
additionally require a causal or off-policy estimator with justified
identification assumptions, including adequate action support.

\paragraph{Precision, optimization bias, and null controls.}
\citet{bokelmann2024uplift} analyze how noisy uplift evaluation affects
precision and how outcome adjustment can reduce required test data.
\citet{cawley2010selection} show that optimizing a noisy model-selection
criterion can itself cause overfitting. \citet{bastani2026falsepromises}
study optimistic model-based policy gains even under seemingly favorable
prediction and splitting conditions; inference-aware optimization incorporates
downstream evaluation power into policy choice~\citep{bastani2025inferenceaware}.
Permutation-based evaluation
provides a reference for performance after disrupting learned dependence
\citep{ojala2010permutation}. In RAISE, the original oracle invariance is an
algebraic consequence of sorting the same rewards used for credit, whereas
refitted perturbations probe a learned selection procedure. A formal
permutation test additionally requires a null-invariant transformation and
consistent recomputation of its full statistic~\citep{hemerik2018exact},
with an appropriate sampling unit and comparison family.

\paragraph{Inspectable LLM signals and collapse diagnostics.}
KAR obtains external knowledge for recommendation~\citep{xi2023kar}, RLMRec
aligns LLM-derived semantic profiles with recommendation representations
\citep{ren2024rlmrec}, and LLMRec augments interaction graphs and profiles
\citep{wei2024llmrec}. Their representation and augmentation settings are
distinct from a per-instance acquisition decision. SHE is an inspectable
signal within this broader design space; its confidence-scored hypothesis
format also connects to HypoGeneAgent in gene-set annotation
\citep{hypogeneagent2025}. RAGEN-2 diagnoses a failure of input-dependent
reasoning during agentic RL~\citep{wang2026ragen2}. In both diagnostic
settings, an aggregate success indicator can leave the operationally
important dependence unmeasured. For RAISE, that dependence is between
before-call context and incremental reward. The empirical evidence comes
from the specified SHE cohorts and reward pipelines; the broader signal
literature supplies further settings in which to investigate this dependence.

\end{document}